\documentclass[10pt,journal,compsoc]{IEEEtran}

\ifCLASSOPTIONcompsoc
  \usepackage[nocompress]{cite}
\else
  \usepackage{cite}
\fi
\ifCLASSINFOpdf
\else
\fi
\usepackage{amsmath, amssymb}
\usepackage{multirow}
\usepackage{mathrsfs}
\usepackage{bm}
\usepackage{color}
\usepackage{graphicx}
\usepackage{makecell}
\usepackage{float}
\definecolor{zzx}{rgb}{0,0,0}

\begin{document}
%
\title{Point Cloud Scene Completion with Joint Color and Semantic Estimation from Single RGB-D Image}
%
%
%
%

\author{Zhaoxuan~Zhang,
	Xiaoguang~Han,
	Bo~Dong,
	Tong~Li,
	Baocai~Yin,
	Xin~Yang
\IEEEcompsocitemizethanks{
	\IEEEcompsocthanksitem Z. Zhang, T. Li, B. Yin and X. Yang are with the School of Computer Science, Dalian University of Technology, Dalian, Liaoning 116024, China. E-mail: zhangzx@mail.dlut.edu.cn, ltx3@mail.dlut.edu.cn,  
	\protect \\ybc@dlut.edu.cn, xinyang@dlut.edu.cn.
\IEEEcompsocthanksitem X. Han is with Shenzhen Research Institute of Big Data, The Chinese University of Hong Kong (ShenZhen), Shenzhen 518172, China. \protect \\
E-mail: hanxiaoguang@cuhk.edu.cn.
\IEEEcompsocthanksitem B. Dong is with Princeton University, Princeton, NJ 08544, USA. \protect \\
E-mail: bo.dong@princeton.edu.
\IEEEcompsocthanksitem X. Yang is the corresponding author.

}
}
\IEEEtitleabstractindextext{%
\begin{abstract}
We present a deep reinforcement learning method of progressive view inpainting for colored semantic point cloud scene completion under volume guidance, achieving high-quality scene reconstruction from only a single RGB-D image with severe occlusion. Our approach is end-to-end, consisting of three modules: 3D scene volume reconstruction, 2D RGB-D and segmentation image inpainting, and multi-view selection for completion. Given a single RGB-D image, our method first predicts its semantic segmentation map and goes through the 3D volume branch to obtain a volumetric scene reconstruction as a guide to the next view inpainting step, which attempts to make up the missing information; the third step involves projecting the volume under the same view of the input, concatenating them to complete the current view RGB-D and segmentation map, and integrating all RGB-D and segmentation maps into the point cloud. Since the occluded areas are unavailable, we resort to a A3C network to glance around and pick the next best view for large hole completion progressively until a scene is adequately reconstructed while guaranteeing validity. All steps are learned jointly to achieve robust and consistent results. We perform qualitative and quantitative evaluations with extensive experiments on the 3D-FUTURE data, obtaining better results than state-of-the-arts.
\end{abstract}

}

\maketitle

\IEEEdisplaynontitleabstractindextext

%
\IEEEpeerreviewmaketitle

\ifCLASSOPTIONcompsoc
\IEEEraisesectionheading{\section{Introduction}\label{sec:introduction}}
\else
\section{Introduction}
\label{sec:introduction}
\fi

%
%
%
%
\IEEEPARstart{A}{}
3D scene with plentiful color and semantic segmentation information can bring human an immersive virtual reality or augmented reality experience, and it also enables the robotics to perform indoor obstacle avoidance, navigation and other tasks well-founded. 
\begin{figure}[ht]
	\centering
	\includegraphics[width=0.45\textwidth]{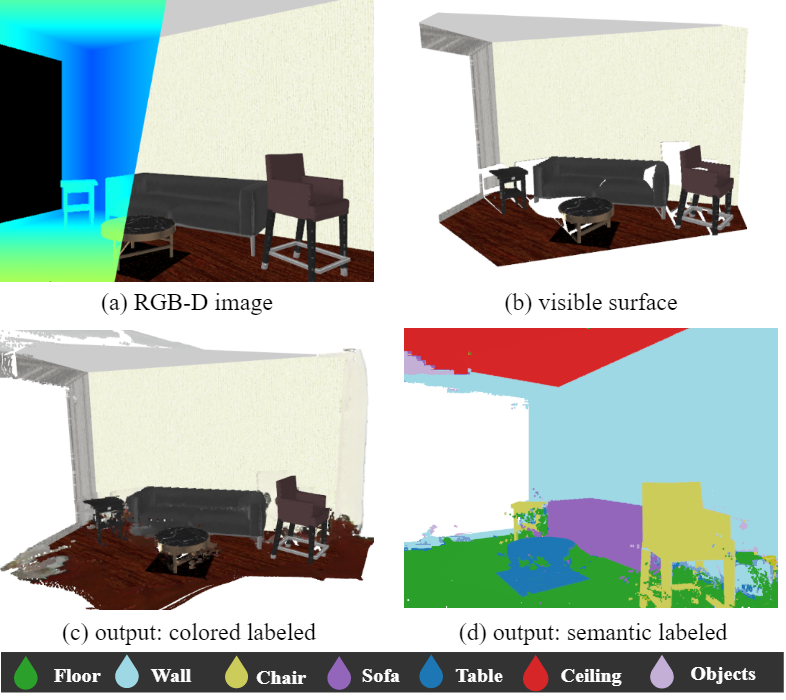}
	\caption{Surface-generated Colored Semantic Point Cloud Scene Completion. (a) A single-view RGB-D image as input; (b) Visible surface from the RGB-D image, which is represented as the point cloud. In our paper, the color of depth map is for visualization only; (c) Our scene completion result with color label: directly recovering the missing points of the occluded regions; (d) Another view of our result labeled with segmentation.}
	\label{fig:teaser}
\vspace{-0.7cm}
\end{figure}

	Thus, reconstructing the 3D scenes from multiple or single images has always been a focus of community's work. Recent years, with the development of deep learning neural networks, previous works\cite{song2017semantic,zhang2018semantic,guo2018view,liu2018see,garbade2019two,wang2018adversarial} focus on reconstructing and completing a 3D voxel scene model from only a single depth or RGB-D image, and predict every voxels' semantic label at the same time, which is also called semantic scene completion(SSC) task. However, the output of SSC is always of volumetric expression, which is a low-resolution representation compared to point cloud. And even though the input of SSC is sometimes an RGB-D image\cite{garbade2019two}, there is still very little work to predict the color information of the model.
	With this motivation, our goal is to implement an algorithm that completes the missing areas of a 3D point cloud scene as well as predicting its colored and semantic segmentation label from a single RGB-D image which we denote as \textbf{colored semantic scene completion}(CSSC) in this paper. {\color{zzx}These missing areas are always not directly captured by the sensor due to object occlusion or self-occlusion.} Although this problem is mild in human vision system, it becomes severe in machine vision because of the sheer imbalance between input and output information. Due to the disorder of point clouds, there is almost no deep learning network that can generate scene-level point clouds, let alone predict the semantic and color information while predicting point coordinates.

Consider the difficulty and complexity of CSSC task, we propose to split it into a sequence of sub-tasks and solving them in turn:
	we first try to recover the missing area for the initial incomplete point cloud from a single depth image, denoted as geometric scene completion(GSC), which has been done in our previous CVPR work\cite{han2019oral}; then we introduce the RGB image corresponding to the input depth map into the algorithm to restore the color information of the entire scene, denoted as colored scene completion(CSC); inspired by \cite{ren2019structureflow}, considering that the semantic information will help both geometry and color completion task, we finally add a semantic segmentation prediction and completion module to the algorithm as an additional guided input for the other two completion tasks, which corresponds to CSSC task.

For the first sub-task, one class of popular approaches \cite{Shao:2012:IAS:2366145.2366155, Chen:2014:ASM:2661229.2661239, GuptaAGM15, guo2015predicting} to this problem is based on classify-and-search: pixels of the depth map are classified into several semantic object regions, which are mapped to most similar 3D ones in a prepared dataset to construct a fully 3D scene. Owing to the limited capacity of the database, results from classify-and-search are often far away from the ground truth. By transforming the depth map into an incomplete point cloud, Song et al. \cite{song2017semantic} recently presented the first end-to-end deep network to map it to a fully voxelized scene, while simultaneously outputting the class labels each voxel belongs to. The availability of volumetric representations makes it possible to leverage 3D convolutional neural networks (3DCNN) to effectively capture the global contextual information, however, starting with an incomplete point cloud results in loss of input information and consequently low-resolution outputs. Several recent works \cite{liu2018see,guo2018view,garbade2019two,wang2018adversarial} attempt to compensate the lost information by extracting features from the 2D input domain in parallel and feeding them to the 3DCNN stream. 
For the second sub-task, less previous work directly address this issue. Wang et al. \cite{wang2020pixel2mesh} propose an end-to-end method for mesh object reconstruction from a single RGB image, which can produce more accurate 3D shape and also predict per-vertex properties(e.g. color). As for scene-level color estimation, the problem become more difficult due to multiple different objects in the scene that block each other. At this time, the information provided by the input 2D image is very limited for colored scene completion task. For the third sub-task, although several recent works \cite{song2017semantic, liu2018see, guo2018view, garbade2019two, wang2018adversarial} have done the semantic information prediction task at the same time with the scene completion task, but they always regard the semantic label of voxel as a multitask output and less consider whether a pre-predicted semantic label will help scene completion task. Following \cite{ren2019structureflow}, we add the semantic segmentation image as the local structure information to guide the RGB-D image inpainting, where we argue better local information should help to improve inpainting quality further then the global context provided by low-resolution voxels.
To our best knowledge, no work has been done on addressing the low-resolution issue of improving output quality and predicting the both color and segmentation label at the same time.


Taking an RGB-D image as input, in this work we advocate the approach of straightforwardly reconstructing 3D points to fill missing region and estimate both color and semantic labels at the same time to achieve high-resolution colored semantic completion (Fig. \ref{fig:teaser}). To this end, we propose to carry out semantic segmentation prediction on initial RGB-D input and completion on multi-view RGB-D and segmentation images in an iterative fashion until all holes are filled, with each iteration focusing on one viewpoint. At each iteration/viewpoint, we render both an RGB-D image and a segmentation map relative to the current view and fill the produced holes using 2D inpainting. The recovered pixels are re-projected to 3D color and semantic labeled points and used for the next iteration. Our approach has two issues: First, different choices of sequences of viewpoints strongly affect the quality of final results because given a partial point cloud, different visible contexts captured from myriad perspectives present various levels of difficulties in the completion task, producing diverse prediction accuracies; moreover, selecting a larger number of views for the sake of easier inpainting to fill smaller holes in each iteration will lead to error accumulation in the end. Thus we need a policy to determine the next best view as well as the appropriate number of selected viewpoints. Second, although existing deep learning based approaches \cite{pathak2016context,iizuka2017globally,liu2018image, ren2019structureflow} show excellent performance for image completion, directly applying them to RGB-D and segmentation maps across different viewpoints usually yields inaccurate and inconsistent reconstructions. The reason is because of lack of global and local context understanding.
To address the first issue, we employ a reinforcement learning optimization strategy for view path planning. In particular, the current state is defined as the updated point cloud after the previous iteration and the action space is spanned by a set of pre-sampled viewpoints chosen to maximize 3D content recovery. The policy that maps the current state to the next action is approximated by a multi-view convolutional neural network (MVCNN) \cite{su15mvcnn} for classification. The second issue is handled by a volume-guided view completion deepnet. It combines the 2D inpainting network \cite{liu2018image, ren2019structureflow} and another 3D completion network \cite{song2017semantic} to form a joint learning machine. In it low-resolution volumetric results of the 3D net are projected and concatenated to inputs of the 2D net, lending better global context information to segmentation image inpainting, and the completed segmentation map will lead better local structure information to RGB-D image inpainting. At the same time, losses from the 2D net are back-propagated to the 3D stream to benefit its optimization and further help improve the quality of 2D outputs. As demonstrated in our experimental results, the proposed joint learning machine significantly outperforms existing methods quantitatively and qualitatively.

\subsection{Contributions}

Existing methods of deep learning semantic scene completion from single (RGB-)D image are always given scene models with volumetric representation which is of low resolution. Our previous work \cite{han2019oral} offers a preliminary attempt at outputting high resolution results, point cloud, when given a single depth map, by adopting an end-to-end volume-guided progressive view inpainting method. Initially, our goal is to reconstruct a complete 3D point cloud scene only from a single RGB-D image, and try to understand the scene as much as possible, such as recovering its color and semantic information. After completing the prediction of the geometric information, in this presented paper, we begin to estimate the color and semantic information of the generated points by inputting RGB image. Inspired by \cite{ren2019structureflow}, considering that the semantic information will help both geometry and completion completion task, we design to use the semantic information of the point cloud as an intermediate result rather than the final output of the network. In particular, we add an RGB image inpainting branch to the volume-guided image inpainting module, and a semantic segmentation prediction and completion module to the algorithm as an additional guided input for both geometry and color completion task. Also, we replace the reinforcement learning algorithm from DQN to A3C for better storage efficiency and more robust performance. In summary, our contributions are as follows.
\begin{itemize}
  \item The first surface-generated algorithm for colored semantic point cloud scene completion from a single RGB-D image by directly predicting the coordinates of the missing points (scene geometry information) and its colored and semantic labels. Both quantitative and qualitative results show the effectiveness of color and semantic information for helping point cloud geometric scene completion task.
  \item A novel deep reinforcement learning strategy for determining the optimal sequence of viewpoints for progressive colored semantic scene completion. Ablation studies prove the effectiveness of the reward functions and the superiority of the view path planning strategy itself compared to choosing viewpoints in an orderly manner.
  \item A volume-guided view inpainting module that not only produces high-resolution outputs but also makes full use of the global context. The projected volumetric segmentation and depth map improve the performance of both inpainting network and volume completion network by end-to-end training procedure.
\end{itemize}

\section{Related Works}
Many prior works are related to scene completion. The literature review is conducted in the following aspects.

\begin{figure*}
	\centering
	\includegraphics[width=0.975\textwidth]{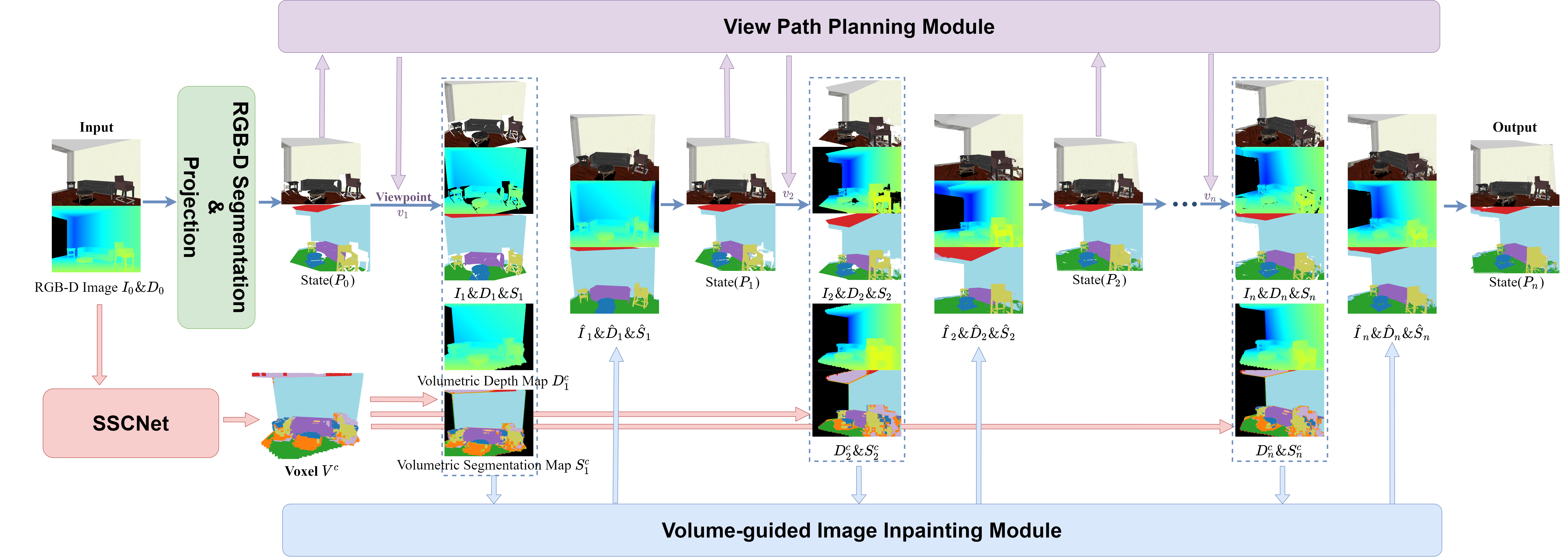}
	\caption{The pipeline of our method. Given a single RGB-D image $I_0 \& D_0$, we first predict its segmentation map $ S_0 $, then we convert $ I_0,D_0,S_0 $ to a colored semantic labeled point cloud $P_0$, here shown in two different views. View path planning module is used to seek the next-best-view $ v_1 $, under which the point cloud is projected to a new RGB-D image $I_1 \& D_1$ and a new segmentation image $ S_1 $, causing holes. In parallel, the $P_0$ is also completed in volumetric space by SSCNet, resulting in $V^c$. Under the viewpoint $ v_1 $, $V^c$ is projected to $ D_1^c, S_1^c $ using for guiding the inpainting of $I_1, D_1, S_1$ with image inpainting module. Repeating this process several times, we can achieve the final high-quality colored semantic scene completion.}
	\label{fig:pipeline}
\vspace{-0.6cm}
\end{figure*}

\textbf{Geometry Completion}
Geometry completion has a long history in 3D processing, known for cleaning up broken single objects or incomplete scenes. Small holes can be filled by primitives fitting\cite{schnabel2009completion, li2011globfit}, smoothness minimization\cite{sorkine2004least, zhao2007robust, kazhdan2013screened}, or structures analysis\cite{mitra2006partial, sipiran2014approximate, sung2015data}. These methods however seriously depend on prior knowledge. Template or part based approaches can successfully recover the underlying structures of a partial input by retrieving the most similar shape from a database, matching with the input, deforming disparate parts and assembling them\cite{shen2012structure, kim2013learning, rock2015completing, sung2015data}. However, these methods require manually segmented data, and tend to fail when the input does not match well with the template due to the limited capacity of the database. Recently, deep learning based methods have gained much attentions for shape completion\cite{rock2015completing, thanh2016field, sharma2016vconv, varley2017shape, dai2017shape, han2017high}, while scene completion from sparse observed views remains challenging due to large-scale data loss in occluded regions. Song et al.\cite{song2017semantic} first propose an end-to-end network based on 3DCNNs, named SSCNet, which takes a single depth image as input and simultaneously outputs occupancy and semantic labels for all voxels in the camera view frustum. ScanComplete\cite{dai2018scancomplete} extends it to handle larger scenes with varying spatial extent. Wang et al.\cite{wang2018adversarial} combine it with an adversarial mechanism to make the results more plausible, and propose a novel architecture in 2019 \cite{wang2019forknet}, named ForkNet, that leverages a shared embedding encoding both geometric and semantic surface cues, as well as multiple generators designed to deal with limited paired data. Zhang et al.\cite{zhang2018semantic} apply a dense CRF model followed with SSCNet to further increase the accuracy. In order to exploit the information of input images, Garbade et al.\cite{garbade2019two} adopt a two stream neural network, leveraging both depth information and semantic context features extracted from the RGB images. Guo et al.\cite{guo2018view} present a view-volume CNN which extracts detailed geometric features from the 2D depth image and projects them into a 3D volume to assist completed scene inference. Li et al. \cite{li2020anisotropic} propose a novel anisotropic convolution module using in 3D convolution network which can handle the object variations in the semantic scene completion from an RGB-D image. Chen et al. \cite{chen20203d} propose a novel 3D sketch-aware feature embedding scheme which explicitly embeds geometric information with structure-preserving details. However, all these works based on the volumetric representation result in low-resolution outputs. In this paper, we directly predict point cloud to achieve high-resolution completion by conducting inpainting on multi-view 2D images.

\textbf{Image Inpainting}
Similar to geometry completion, researchers have employed various priors or optimized models to complete a depth image\cite{herrera2013depth, liu2012guided, muddala2014depth, thabet20143d, chen2014improved, liu2016building, xue2017depth, zhang2018probability}. The patch-based image synthesis idea is also applied\cite{doria2012filling, gautier2011depth}. Recently, significant progresses have been achieved in image inpainting field with deep convolutional networks and generative adversarial networks (GANs) for regular or free-form holes\cite{iizuka2017globally, liu2018image, yu2018free, li2020inpaint, lahiri2020prior}. Zhang et al.\cite{zhang2018deep} imitate them with a deep end-to-end model for depth inpainting. Compared with inpainting task on irregular holes , recovering missing information in our task is more challenging due to the holes are always created by mutual occlusion of objects. Shih et al. \cite{shih20203d} present a learning-based inpainting model that synthesizes new local color-and-depth content into the occluded region in a spatial context-aware manner, which is similar to our work. But they only inpaint the holes guided by 2D context features, which may lead some unreasonable results.
To address it, an additional 3D global context is provided in our paper, guiding the inpainting on diverse views to reach more accurate and consistent output.

\textbf{Deep Learning On Point Cloud}
Deep learning has been introduced to various point cloud processing tasks to improve performance of algorithms, such as classification, segmentation, object generation and completion. Our CSSC task belongs to point cloud scene completion and segmentation. 
Achlioptas et al.\cite{achlioptas2017learning} introduce the first deep generative model for the point cloud. Due to the model architecture is not primarily built to do geometry completion task, the completion performance is not considered ideal.  Yuan et al.~\cite{yuan2018pcn} propose the first learning-based architecture, named point completion network (PCN), focusing on shape completion task. PCN applies the Folding operation ~\cite{yang2018foldingnet} to approximate a relatively smooth surface and conduct shape completion. Recently, Sarmad et al.~\cite{sarmad2019rl} propose a reinforcement learning agent controlled GAN based network (RL-GAN-Net) for real-time point cloud completion. The RL agent used in RL-GAN-Net avoids complex optimization and accelerates the prediction process, and it does not focus on enhancing the accuracy of the points. Huang et al.~\cite{huang2020pf} propose a learning-based approach, point fractal network (PF-Net), for precise and high-fidelity point cloud completion. PF-Net preserves the spatial arrangements of the incomplete point cloud and can predict the detailed geometrical structure of the missing region(s) in the prediction. However, all these works only solve the object-level point completion task, but not for the larger scale situation. To address it, we project the large scale point clouds to multi-view 2D images and achieve the point-cloud completion by image inpainting.

The semantic segmentation on point clouds is the expansion of 2D-Image domains. Huang et al.~\cite{huang2016point} first propose 3D fully convolutional neural networks(3D-FCNN) which predicts coarse voxel-level semantic label. PointNet~\cite{qi2017pointnet} and following works~\cite{qi2017pointnet++,engelmann2017exploring} use multi-layer perception (MLP) to predict higher resolution point-level segmentation results. Pham et al.~\cite{pham2019jsis3d} propose a semantic-instance segmentation method that jointly performs both of the tasks via a novel multi-task pointwise network and a multi-value conditional random field model. Wang et al.~\cite{wang2019graph} design a novel graph attention convolution (GAC) framework with learnable kernel shape for structured feature learning of 3D point cloud and apply it to train an end-to-end network for semantic segmentation task. Jiang et al.~\cite{jiang2019hierarchical} explore semantic relation between each point and its contextual neighbors through edges, and propose a hierarchical point-edge interaction network with point branch as well as edge branch.  All these works take the point cloud as the input and design different networks to treat the 3D points directly. However, our approach only takes a single RGB-D image and outputs a completed and colored semantic labeled point cloud.

\textbf{View Path Planing}
Projecting a scene or an object to the image plane will severely cause information loss because of self-occusions. A straightforward solution is utilizing dense views for making up\cite{su15mvcnn, qi2016volumetric, tatarchenko2016multi}, yet it will lead to heavy computation cost. Choy et al.\cite{choy20163d} propose a 3D recurrent neural networks to integrate information from multi-views which decreases the number of views to five or less. Even so, how many views are sufficient for completion and which views are better to provide the most informative features, are still open questions. Optimal view path planning, as the problem to predict next best view from current state, has been studied in recent years. It plays critical roles for scene reconstruction as well as environment navigation in autonomous robotics system\cite{low2006adaptive, blaer2007data, zhou2013dense, wu2014quality}. Most recently, this problem is also explored in the area of object-level shape reconstruction\cite{yang2018active}. A learning framework is designed in \cite{xu20163d}, by exploiting the spatial and temporal structure of the sequential observations, to predict a view sequence for groundtruth fitting. Our work explores the approaches of view path planning for scene completion. 
We propose to train an asynchronous advantage actor-critic (A3C) algorithm \cite{mnih2016asynchronous} to choose the best view sequence in a reinforcement learning framework.
\section{Methodology}

\begin{figure}
	\centering
	\includegraphics[width=0.475\textwidth]{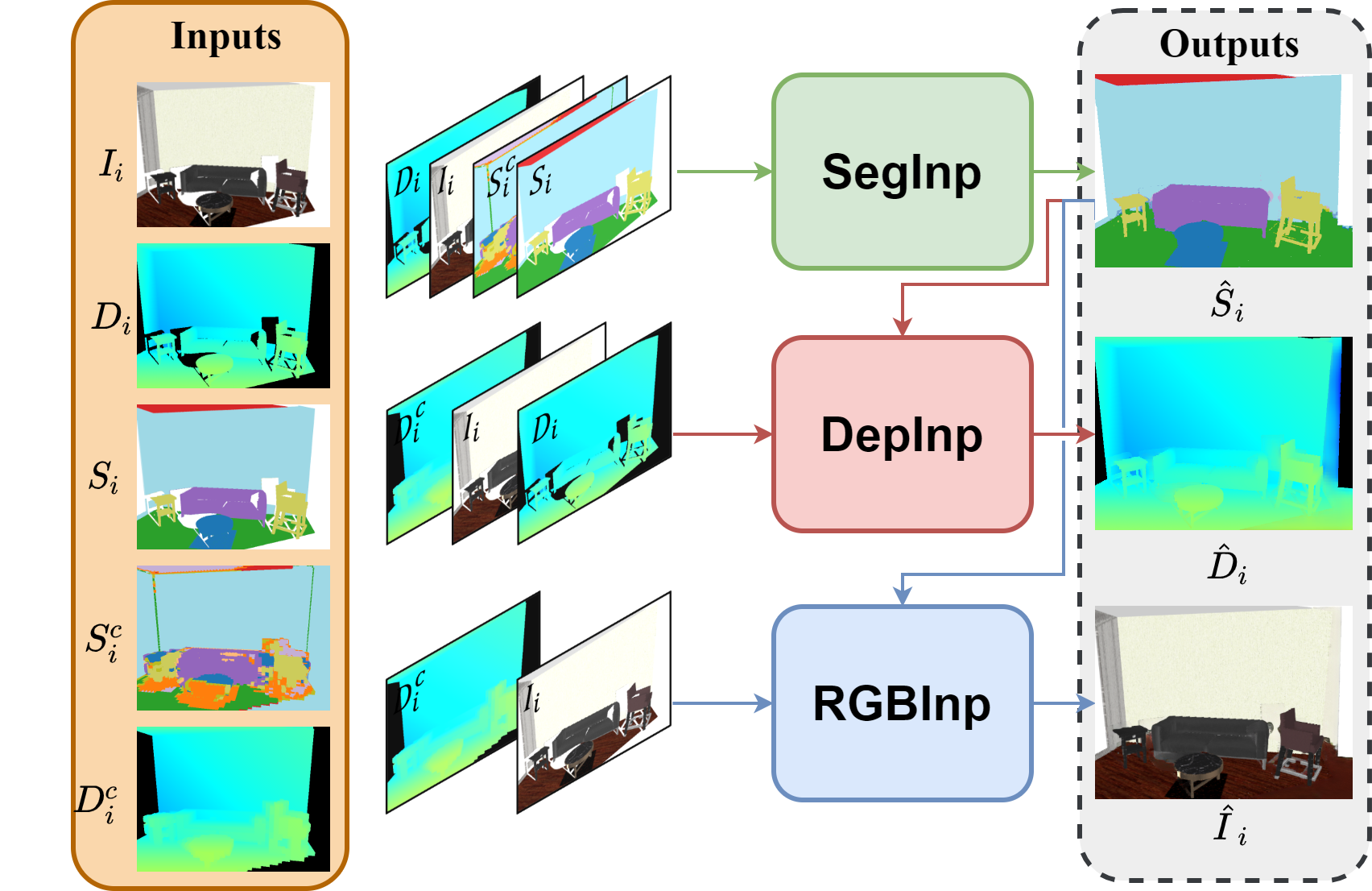}
	\caption{Details of our volume-guided image inpainting module. The output of segmentation map inpainting network $ \hat{S_i} $ will be input to both depth inpainting network and RGB inpainting network, as it can provide structure information.}
	\label{fig:inpmodule}
\vspace{-0.5cm}
\end{figure}

\subsection{Overview}
The insight behind our approach is that viewing an incomplete 3D point cloud from different viewpoints should give us more information about the incompleteness than viewing it under a fixed
angle. Therefore, an effective way to fix the incompleteness is sequentially fixing the observed holes under different viewpoints until we cannot obverse any hole. Our main idea is projecting an
incomplete point cloud into multi-view 2D images(include RGB image, depth map and segmentation map) and performing 2D inpainting tasks on them. Then, the restored 2D images are converted back to the 3D point cloud to obtain a complete colored semantic point cloud. In order to ensure 2D image inpainting task have sufficient global and local information, we introduce the 3D volumetric occupancy to provide valuable global information, and we propose to leverage semantic segmentation on RGB-D images to provide local information. 

In particular, taking an RGB-D image $I_0 \& D_0$ as input, we first predict the semantic label for each pixel, which results in a semantic segmentation map $ S_{0} $, and convert $ I_0, D_0, S_0 $ to a colored semantic labeled (CSL) point cloud $P_0^{CSL}$, which suffers from severe data loss.
Our goal is to generate 3D points to complete $P_0^{CSL}$'s lack of geometry information, and predict the inferred points' color and segmentation information at the same time. To take full advantage of the context information, we execute the inpainting operations view by view in an accumulative way, with inferred points for the current viewpoint kept and used to help inpainting of the next viewpoint. Assume $ I_0, D_0, S_0 $ is rendered from $P_0^{CSL}$ under viewpoint $v_0$, we start our completion procedure with a new view $v_1$ and render $P_0^{CSL}$ under $v_1$ to obtain three new images $I_1, D_1, S_1$, which potentially have many holes. We fill these holes  with 2D inpainting, turning $I_1, D_1, S_1$ to $\hat{I}_{1}, \hat{D}_{1}, \hat{S}_{1}$. The inferred depth pixels in $\hat{D_1}$ are then converted to 3D points, attached color information and segmentation label from $ \hat{I}_{1}, \hat{S}_{1} $, and aggregated with $P_0^{CSL}$ to output a denser point cloud $P_1^{CSL}$. This procedure is repeated for a sequence of new viewpoints $v_2,v_3,...,v_n$, yielding point clouds $P_2^{CSL},P_3^{CSL},...,P_n^{CSL}$, with $P_n^{CSL}$ being our final output. Fig. \ref{fig:pipeline} depicts the overall pipeline of our proposed algorithm.
Since $P_n^{CSL}$ depends on the view path $v_2,v_3,...,v_n$, we describe in section \ref{sec:a3c} a deep reinforcement learning framework to seek the best view path. Before that, we introduce our solution to another critical problem of 2D inpainting, i.e., transforming $I_i, D_i, S_i$ to $\hat{I}_{i}, \hat{D}_{i}, \hat{S}_{i}$, in section \ref{sec:inpaint} first. 

\subsection{Volume-guided Image Inpainting}
\label{sec:inpaint}

Deep Convolutional Neural Network (CNN) has been widely utilized to effectively extract context features for image inpainting tasks, achieving excellent performance. Although it can be directly applied to each viewpoint independently, this simplistic approach will lead to inconsistencies across views because of lack of global context understandings. We propose a volume-guided view inpainting framework by first conducting completion in the voxel space, converting point cloud $P$'s volumetric occupancy grid $V$ to its semantic labeled completed version $V^c$. Denote the projected depth map from $V^c$ to the view $v_i$ as $D^c_i$, segmentation map from $ V^c $ as $ S^c_i $. As shown in Fig. \ref{fig:pipeline} and Fig. \ref{fig:inpmodule}, this is implemented using a five-module neural network architecture consisting of a volume completion network, a segmentation inpainting network, an RGB inpainting network, a depth inpainting network, and a differentiate projection layer connecting them. In addition, there is a semantic segmentation network before doing image inpainting. The details of each module and our training strategy are described below.

\noindent \textbf{RGB-D Semantic Segmentation}
In our implementation, DANet \cite{fu2019dual} is used as our RGB-D semantic segmentation network, which takes $ 640\times480 $  $ I_0 \& D_0 $ and outputs the same size $ S_0 $.

\noindent \textbf{Volume Completion}
We employ SSCNet proposed in \cite{song2017semantic} to map $V$ to $V^c$ for volume completion. SSCNet predicts not only volumetric occupancy but also the semantic labels for each voxel. Such a multi-task learning scheme helps us better capture object-aware context features and contributes to higher accuracy. The readers are referred to \cite{song2017semantic} for details on how to set up this network architecture. We retain the multi-task output as \cite{song2017semantic}, where the resolution of input is $240\times144\times240$, and the output is $60\times36\times60\times12$ indicating the probability of the grid corresponding to 12 labels (excluding \textit{empty} label).

\noindent\textbf{Segmentation and RGB-D Inpainting} 

	As we have predicted the volume completion $ V^c $, we need to use this global information to better inpaint the segmentation and RGB-D images rendered in different viewpoints, so as to gradually complete the point cloud scene $ P_{0}^{CSL} $.
	In particular, we first complete the holes in segmentation map $ S_i $, and then using it to guide the inpainting of RGB-D images $ I_i \& D_i $. Among various existing approaches, PartialCNN \cite{liu2018image} is chosen to handle our case with holes of irregular shapes for segmentation and depth inpainting, and StructureFlow \cite{ren2019structureflow} is chosen to better predict the color information through flow for RGB inpainting. Specifically, for segmentation inpainting, $S_{i}, I_i, D_i$ and $S^{c}_{i}$ are first concatenated to form a map. The resulting map is then fed into a U-Net structure implemented with a masked and re-normalized convolution operation (also called partial convolution), followed by an automatic mask-updating step. The output $ \hat{S_i} $ is also in $640\times480$. Identically, for depth inpainting, $ D_{i}, I_{i}, D^{c}_{i} $ and the completed segmentation map $ \hat{S_i} $ are concatenated as the input, and $ \hat{D_i} $ is the output in $ 640\times480 $. As for RGB inpainting, $ I_{i}, D_{i}^{c}, \hat{S_i} $ are concatenated as a map to feed into a U-Net color generator and also into a encoder to generate appearance flow. Guided by the flow information, colors are predicted from regions with similar structures. The output $ \hat{I_i} $ is also with size $ 640\times480 $. We refer the readers to \cite{liu2018image,ren2019structureflow} for details of the architecture settings and the design of loss functions. 


\noindent\textbf{Projection Layer} As validated in our experiments described in section \ref{abs:img}, the projection of $V^c$ greatly benefits inpainting of 2D maps.  We further exploit the benefit of 2D inpainting to volume completion by propagating the 2D loss back to optimize the parameters of 3D CNNs. Doing so requires a differentiable projection layer.
There are two options for the implementation of this layer: the technique proposed in \cite{tulsiani2017multi} and the homography warping method in \cite{yao2018mvsnet}. The first one is chosen for a more accurate projection.
Thus, we connect $V^c$ and $D^c_i, S^c_i$ using this layer. For the sake of notational convenience, we use $V$ to represent $V^c$ ,$D$ to represent $D^c_i$ and $ S $ to represent $ S^c_i $. Specifically, for each pixel $x$ in $D$ or $ S $, we launch a ray that starts from the viewpoint $v_i$, passes through $x$, and intersects a sequence of voxels in $V$, noted as $l_{1}, l_{2},..., l_{N_{x}}$. We denote the value of the $k_{th}$ voxel's first channel in $V$ as $V_k$, which represents the probability of this voxel belong to empty label, and denote the value of other channels as $ s_k $. Then, we define the depth value of this pixel $x$ as
\begin{equation}\label{key}
D(x)=\sum^{N_{x}}_{k=1}P^{x}_{k} d_{k},
\end{equation}
where $d_{k}$ is the distance from the viewpoint to voxel $l_{k}$ and $P^{x}_{k}$ the probability of the ray corresponding to $x$ first meets the $l_{k}$ voxel
\begin{equation}\label{key}
P^{x}_{k}=(1-V_{k})\prod^{k-1}_{j=1}V_{j},  \ k=1,2,...,N_{x}.
\end{equation}

The segmentation value of the pixel $ x $ as
\begin{equation}\label{key}
S(x) = \sum^{N_{x}}_{k = 1}P^{x}_{k} \cdot soft\arg \max(s_{k}).
\end{equation}

The derivative of $ D(x)$ with respect to $V_{k}$ can be calculated as
\begin{equation}\label{key}
\frac{\partial D(x)}{\partial V_{k}}=\sum_{i=k}^{N_{x}}(d_{i+1}-d_{i})\prod_{1\le t\le i, t\neq k}V_{t}.
\end{equation}

And the derivative of $ S(x)$ with respect to $V_{k}$ can be calculated as
\begin{equation}\label{key}
\frac{\partial S(x)}{\partial V_{k}}= \sum_{i=k}^{N_{x}}(u_{i+1}-u_{i})\prod_{1\le t\le i, t\neq k}V_{t} \cdot \frac{\partial u(\cdot)}{\partial s(\cdot)},
\end{equation}

where $ u_i = soft\arg \max(s_{i}) $. The derivative of $ S(x)$ with respect to $s_{k}$ can be calculated as
\begin{equation}\label{key}
\frac{\partial S(x)}{\partial s_{k}}= P^{x}_{k} \cdot \frac{\partial u(\cdot)}{\partial s(\cdot)}.
\end{equation}

This guarantees back propagation of the projection layer. In order to speed up implementation, the processing of all rays are implemented in parallel via GPUs.

\noindent \textbf{Joint Training} Training the five components together from scratch is tricky, and training convergence and stability are not guaranteed. Instead, the proposed inpainting network is trained as follows: 1) We pre-train both the RGB-D semantic segmentation network and the volume completion network independently. 2) With fixed parameters of the RGB-D semantic segmentation network and the volume completion network, we train the segmentation inpainting network. 3) We train depth inpainting network and RGB inpainting network with fixed parameters of the other three networks independently. 4) Once we have all pre-trained models for the five networks, we train the entire network jointly.

The training data are generated based on the 3D-FUTURE synthetic scene dataset provided in \cite{fu20203dfuture}. {\color{zzx}Please go Sec. \ref{sce:exp} for more generation details.}

\subsection{Progressive Scene Completion}
\label{sec:a3c}
Given an incomplete point cloud $P_0^{CSL}$ that is converted from $I_0,D_0,S_0$ with respect to view $v_0$, we describe in this subsection how to obtain the optimal next view sequence $v_1, v_2,...,v_n$.
The problem is defined as a Markov decision process (MDP) consisting of state, action, reward, and an agent which takes actions during the process. The agent inputs the current state, outputs the corresponding optimal action, and receives the most reward from the environment. We train our agent using A3C \cite{mnih2016asynchronous}, an algorithm of deep reinforcement learning. We adopt the A3C as our viewpoint sequence planning method for the reason that off-policy reinforcement learning methods like DQN \cite{mnih2015human} take up additional memory space caused by the experience replay, and at the same time A3C can achieve faster training speed and better viewpoint path selection because of the more diverse data brought by multi-agent parallelism. The ablation study results prove that the A3C is more suitable for our complex tasks than DQN. The definitions of the proposed MDP and the training procedure are given below.

\noindent \textbf{State} We define the state as the updated point cloud at each iteration, with the initial state being $P_0^{CSL}$. As the iteration continues, the state for performing completion on the $i_{th}$ view is $P_{i-1}^{CSL}$, which is accumulated from all previous iteration updates.

\begin{figure*}[ht]
	\centering
	\includegraphics[width=0.95\textwidth]{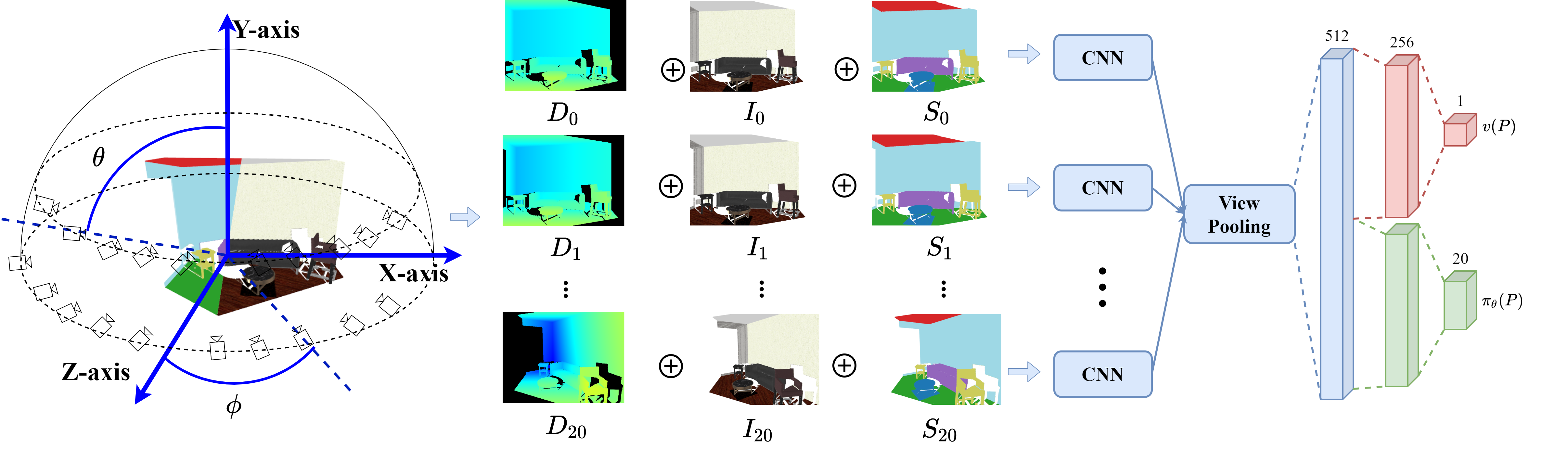}
	\caption{The architecture of a single agent of our A3C. For a point cloud state, MVCNN is used to predict the best view for the next inpainting.}
	\label{fig:A3C}
\end{figure*}

\noindent \textbf{Action Space} The action at the $i_{th}$ iteration is to determine the next best view $v_i$. To ease the training process and support the use of A3C, we evenly sample a set of scene-centric camera views to form a discrete action space. Specifically, we first place $P_0^{CSL}$ in its bounding sphere and keep it upright. {\color{zzx}At this point, $ v_0 $ is located on $ z $-$ y $ plane, and camera points to the coordinate system origin.} Then, two circle paths are created for both the equatorial and 70-degree latitude line. In our experiments, $20$ camera views are uniformly selected on these two paths, $10$ per circle. All views are facing to the center of the bounding sphere. We fixed these views for all training samples. The set of 20 views is denoted as $C = \{c_1,c_2,...,c_{20}\}$.
In particular, a viewpoint is defined as:

\begin{equation}\label{key}
\begin{array}{c}
x = a\sin\theta\sin\phi, \\
y = a\cos\theta, \\
z = a\sin\theta\cos\phi,
\end{array}
\end{equation}
where $ \theta \in \{ 70^{\circ},90^{\circ} \} $ and $ \phi \in \{-50^{\circ},-40^{\circ},-30^{\circ},-20^{\circ}, \-10^{\circ},
10^{\circ}, 20^{\circ},30^{\circ},40^{\circ},50^{\circ} \} $. {\color{zzx}At this point, $ [\theta, \phi] $ for $ c_1 $ is $[90^{\circ},-50^{\circ}]$, $[90^{\circ},-40^{\circ}]$ for $ c_2 $, $[70^{\circ},-40^{\circ}]$ for $ c_{11} $ and $[70^{\circ},50^{\circ}]$ for $ c_{20} $.} As shown in Fig.~\ref{fig:A3C}, all viewpoints are facing to the scene center and $ a $ is set to $ 3m$ in our experiments.

\noindent\textbf{Reward} An reward function is commonly unitized to evaluate the result for an action executed by the agent. In our work, at the $i_{th}$ iteration, the inputs are three incomplete maps $I_i, D_i, S_i$ rendered from $P_{i-1}^{CSL}$ under view $v_i$ chosen in the action space $C$. The result of the agent action are corresponding inpainted image $\hat{I_i},\hat{D_i},\hat{S_i}$. Hence the accuracy of this inpainting operation can be used as the primary rewarding strategy. It can be measured by the mean error of the pixels inside the holes between $\hat{I_i},\hat{D_i},\hat{S_i}$ and its ground truth $I^{gt}_i,D^{gt}_i,S^{gt}_i$. All the ground truth maps are pre-rendered from 3D-FUTURE dataset. Thus we define the award function as
\begin{equation}\label{equ1}
R_{i}^{acc}=-\frac{1}{3|\Omega|}(L^1_{\Omega}(\hat{D_i}, D^{gt}_i) + L^1_{\Omega}(\hat{I_i}, I^{gt}_i) + L^1_{\Omega}(\hat{S_i}, S^{gt}_i)),
\end{equation}
where $L^1$ denotes the $ L_{1}$ loss, $\Omega$ the set of pixels inside the holes, and $|\Omega|$ the number of pixels inside $\Omega$.

If we only use the above reward function $R_{i}^{acc}$, the agent tends to change the viewpoint slightly in each action cycle, since doing this results in small holes. However, this incurs higher computational cost while accumulating errors. We thus introduce a new reward term to encourage inferring more missing points at each step.
This is implemented by measuring the percentage of filled original holes. To do so, we need to calculate the area of missing regions in an incomplete point cloud $P$, which is not trivial in a 3D space. Therefore, we project $P$ under all camera views to the action space $C$ and count the number of pixels inside the generated holes in each rendered image. The sum of these numbers is denoted as $Area^{h}(P)$ for measuring the area.
We thus define the new reward term as
\begin{equation}\label{equ2}
R_{i}^{hole}=\frac{Area^{h}(P_{i-1})-Area^{h}(P_{i})}{Area^{h}(P_0)}-1
\end{equation}
to avoid the agent from choosing the same action as in previous steps.
We further define a termination criterion to stop view path search by $ Area^{h}(P_{i})/Area^{h}(P_{0}) < 7\%$, which means that all missing points of $P_0$ have been nearly recovered. We set the reward for terminal to 1.

However, the functions $ R_i^{acc} $ and $ R_i^{hole} $ are focused on the quality of the inpainted maps in 2D space only. Therefore, the chosen viewpoint may not directly enforce predicting more accurate 3D points. To this end, we introduce point cloud recover reward function $R^{pcacc}_{i}$, and it is defined as:
\begin{align}
R^{pcacc}_{i} &=  \frac{1}{N}\sum_{j=1}^N f(p_j, P_{GT}) , \ p_j \in \tilde{P}^{CSL}_i, \label{eq:pcacc_def}\\
f(p, P) &= \begin{cases} 1, & L(p) = L(\underset{q \in P }{\arg\min} \| p - q \|_{2}^{2} ) \\ 
0, & \text{others},  \end{cases}
\end{align}
where, $N$ is the total number of points in $\tilde{P}^{CSL}_i$; $\tilde{P}^{CSL}_i$ is a 3D point set and it contains all recovered 3D points from iteration $i$. In other words, $P^{CSL}_i = P^{CSL}_{i-1} + \tilde{P}^{CSL}_i$; $P_{GT}$ is the ground truth point cloud; $L(\cdot)$ represents the segmentation label of a point. Intuitively, given a predicted 3D point $p_j$, we calculate the closest point to $ p_j $ in $ P_{GT} $, marked as $ q $. If the predicted semantic label of $p_j$ (i.e., $L(p_j)$) is the same as the semantic label of the closet point $ q $, we then count $p_j$ as a correct prediction. Otherwise, it is counted as an incorrect prediction. $R^{pcacc}_{i}$ estimates the percentage of the correct predictions among all predictions. Combining all three reward functions, the final reward function is:


\begin{equation}\label{reward}
R^{total}_{i} = \alpha R^{acc}_{i} + \beta (R^{pcacc}_{i} - 1) +  \gamma R^{hole}_{i},
\end{equation}
where $ \alpha, \beta, \gamma $ are the balancing weights.

\noindent \textbf{A3C Training} Our A3C is built upon MVCNN\cite{su15mvcnn}. It consists of $ n $ local networks and a global network, which have same architecture but time difference in network parameters updating. These local networks are trained in parallel to update the parameters of the global network.
It takes mutil-view RGB-D and segmentation maps projected from $ P_{i-1} $ as inputs and outputs the value of input state $ v_{\theta_{v}}(P) $ from critic branch as well as action probability distribution $ \pi _{\theta} (P) $ from actor branch. The whole network is trained to approximate the correct probability $ \pi _{\theta}(v_{i} | P_{i-1}) $ for taking action $ v_{i} $ and the value function $ v_{\theta_{v}}(P_{i-1}) $, which is the expected reward that the agent receives when taking action $ v_{i} $ at state $ P_{i-1} $.

The loss function for training actor is
\begin{equation}\label{key}
Loss(\theta)=\mathbb{E}[\log \pi _{\theta}(v_{i} | P_{i-1})\cdot (R^{total}_{i} + \delta v_{\theta_{v}}(P_{i}) - v_{\theta_{v}}(P_{i-1}))]
\end{equation}

where $ \delta $ is a discount factor. 
The loss function for training critic is
\begin{equation}\label{key}
Loss(\theta_{v})=\mathbb{E}[(R^{total}_{i} + \delta v_{\theta_{v}}(P_{i}) - v_{\theta_{v}}(P_{i-1}))^2]
\end{equation}

Finally, our loss function for training A3C is
\begin{equation}\label{key}
\mathbb{L}_{our} = Loss(\theta) + Loss(\theta_{v}).
\end{equation}

Our network structure is shown in Fig. \ref{fig:A3C}. At state $ P_{i-1} $, we render at all viewpoints $ \{c_{1}, c_{2}, ..., c_{20}\} $ in the action space $ C $ in $ 224\times224 $ resolution and get the corresponding multi-view depth maps $ \{D_{i}^{1}, D_{i}^{2}, ..., D_{i}^{20}\}  $, RGB images $ \{I_{i}^{1}, I_{i}^{2}, ..., I_{i}^{20}\} $ and segmentation maps $ \{S_{i}^{1}, S_{i}^{2}, ..., S_{i}^{20}\} $. These maps are first concatenated according to different viewpoints and then sent to the same $ CNN $ as inputs. After a view pooling layer and a fully-connected layer, we obtain a 512-D vector, which is split evenly into two parts to learn the action probability distribution $ \pi _{\theta} (P) $ and the state value function $ v(P) $. Finally, the agent choose the action $ v_{i} $ for current state $ P_{i-1} $ based on policy $ \pi _{\theta} (P_{i-1}) $. In the end, we reach the decision on maps $ I_i, D_{i}, S_i $ for inpainting.

\section{Experimental Results}
\label{sce:exp}

\noindent \textbf{Dataset}
The dataset we used to train our 2DCNN and A3C is generated from 3D-FUTURE \cite{fu20203dfuture}.{ \color{zzx}3D-FUTURE contains 16,563 unique detailed 3D instances of furniture with high-resolution textures of 5,000 different rooms. Using the camera view generation method provided by \cite{fu20203dfuture}, we generated a total of 2,541 camera viewpoints in 1,000 random rooms and rendered 2,541 RGB-D images from scene models under these camera viewpoints. After filtering the above images and removing the unqualified camera viewpoints as Song \textit{et al.} \cite{song2017semantic}, we finally obtained 1,582 valid camera viewpoints and their corresponding initial RGB-D scene images. Among them, 1,431 images are randomly selected using for training and the remaining 151 for testing. As for ground truth data, we performed frustum cutting of each room model based on those valid camera viewpoints (the distance to the camera is set to $ 0.1m $ and $ 6m $ for the near and far clipping plane, respectively). After generating these local scene mesh model, we sampled around 800,000 points for each model using Poisson Disk Sampling method \cite{bridson2007fast} and then voxelized each point cloud at different resolutions by Voxel Down Sampling algorithm in Open3D library \cite{Zhou2018}. These ground truth point clouds and voxels will be used for the training of the relevant networks and the calculation of the evaluation metrics. Specifically, for 2DCNN, we first re-projected 1,582 RGB-D images into initial point clouds according to their corresponding viewpoints. Then for each initial viewpoint $ v_0 $, we generated 20 viewpoints around it using the similar method as in Sec. \ref{sec:a3c} to avoid causing large holes and to ensure that sufficient contextual information is available in the learning process. After projecting the initial point clouds and ground truth point clouds under these viewpoints, we got $ 1,582 \times 20=31,640 $ pairs of images for training and testing our image inpainting and view path planning networks. In addition, we added noise based on the standard normal distribution to the input depth images, for simulating the image taken by the sensor in the real world.}

{\color{zzx}For the real data, we use the ScanNet V2 dataset \cite{dai2017scannet}, which contains more than 1,500 scanned scenes with cluttered object placement. We randomly selected 100 scenes and calculated 1-3 initial views for each scene according to the complexity of the scene following Song \textit{et al.} \cite{song2017semantic}, and finally obtained 153 real RGB-D scans. The ground truth point cloud and voxel of each local scan are generated in similar way as on the 3D-FUTURE dataset. These data are only used for testing our method and related methods. }

\noindent \textbf{Implementation Details} Our network architecture is implemented in PyTorch. The provided pre-trained model of SSCNet \cite{song2017semantic} is used to initialize parameters of our 3DCNN part. It takes around 60 hours to train the depth inpainting network , RGB inpainting network and segmentation inpainting network on our training dataset separately and 20 hours to fine-tune the whole network after the addition of projection layer. During A3C training process, we set the number $ n $ of local networks to 3, the weight $ \alpha, \beta, \gamma $ for reward calculation to 0.05, 1, 0.1 and the discount factor $ \delta $ to 0.9. {\color{zzx}Compared to 5 days used for training DQN, training A3C only takes 3 days and running our complete algorithm once takes about $ 100 s $ which adopts $ 5.31 $ viewpoints on average.}

\noindent \textbf{Metrics} {\color{zzx}The evaluation metrics will be divided into two aspects: geometric completion and semantic segmentation. The Chamfer Distance (CD) \cite{fan2017point} is used as one of our metrics for evaluate the accuracy of our generated point set $P$, compared with the ground truth point cloud $P_{GT}$. 
 Following Sung \textit{et al.} \cite{sung2015data}, we also use $ C_r(Completeness) $ and $ A_r(Acuraccy) $ to evaluate how complete and accuracy of the generated result. We define it as:}
 \begin{equation}\label{completeness}
 C_r(P, P_{GT}) = \dfrac{\left| \{d(x,P) < r|x\in P_{GT}\}\right|}{ \left| \{y|y\in P_{GT}\}\right|}
 \end{equation}
 
 \begin{equation}\label{accuracy}
A_r(P, P_{GT}) = \dfrac{\left| \{d(x,P_{GT}) < r|x\in P\}\right|}{ \left| \{y|y\in P\}\right|}
 \end{equation}
where $ d(x,P) $ denotes the distance from a point $ x $ to a point set $ P $, $ \left| \cdot \right| $ denotes the number of the elements in the set, and $ r $ means the distance threshold. In our experiments, we report the $ C_r $ and $ A_r $ w.r.t five different $ r $ ($ 0.02,0.04,0.06,0.08,0.10 $ are used).

\begin{figure}[ht]
	\centering
	\includegraphics[width=0.475\textwidth]{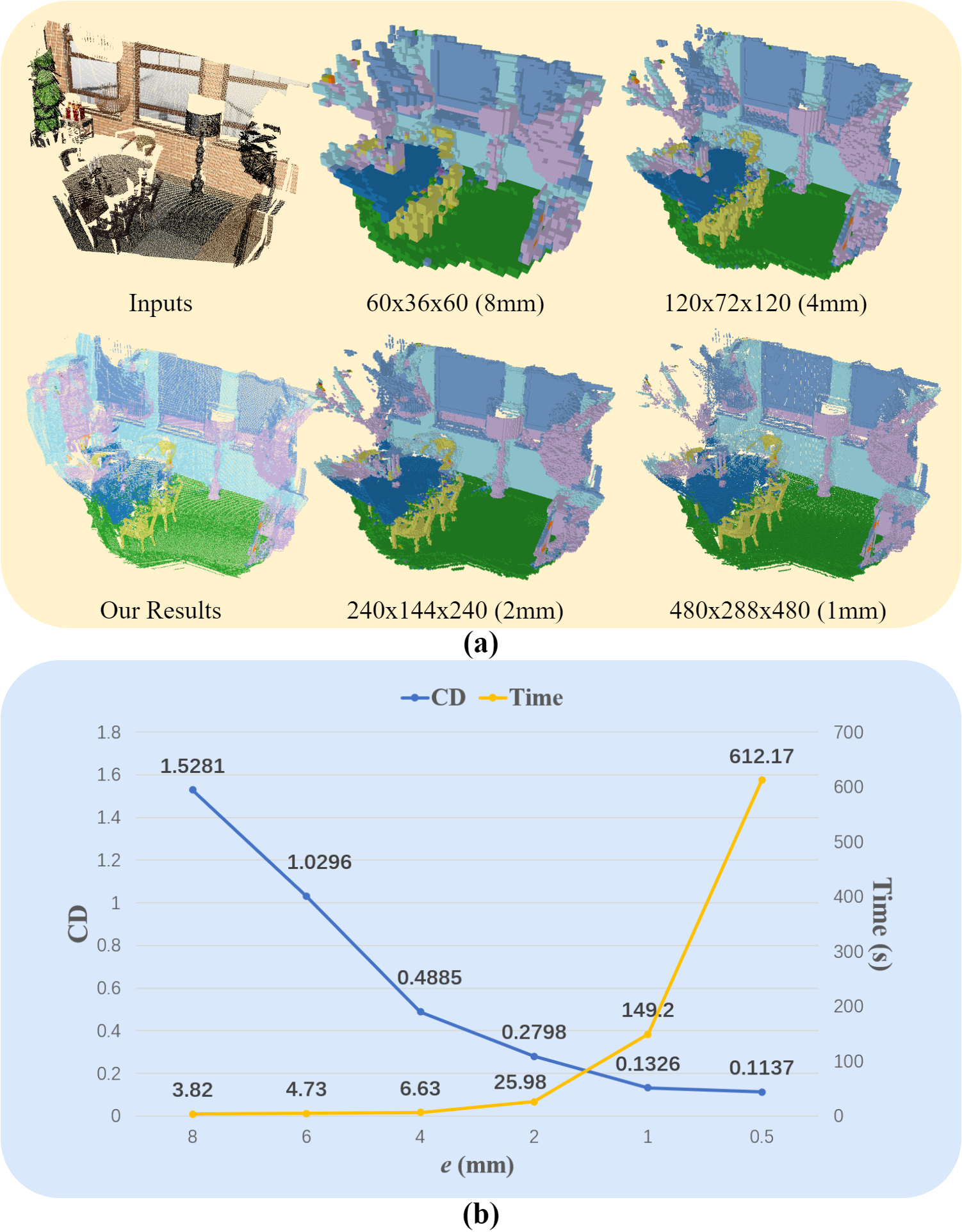}
	\caption{{\color{zzx}(a) Visualization of voxelization results with different resolutions determined by different grid edge lengths. (b) The relationship between different $ e $ on result accuracy (evaluated by CD) and voxelization generating time.} }
	\label{fig:grid}
	\vspace{-0.3cm}
\end{figure}

We also use the same semantic categories as the ones used in~\cite{song2017semantic} for our semantic segmentation validation. The 11 classes are of varying shapes and sizes, and they are: $ \mathcal{L}= \{$\emph{ceiling}, \emph{floor}, \emph{wall}, \emph{window}, \emph{chair}, \emph{bed}, \emph{sofa}, \emph{table}, \emph{tvs}, \emph{furniture}, and \emph{other objects}$ \} $. {\color{zzx}One issue here is that the reconstructed 3D points are not guaranteed to be well aligned with ground truth 3D points. Therefore it is difficult to define True Positive (TP), False Positive (FP) and False Negative (FN) to calculate
Intersection over Union(IoU) directly with $ IoU = TP/(TP+FP+FN) $. For this reason, we choose to voxelize $ P $ and $ P_{GT} $ with different resolution to achieve an one-to-one correspondence between $ P $ and $ P_{GT} $ with minimal loss of accuracy. Specifically, followed by Song \textit{et al.} \cite{song2017semantic}, the size of the local scene model is defined as $ 4.8m \times 2.44m \times 4.8m $. Thus, given different length of voxel grid's edge one can get voxel models of different resolution. Given $ P $ and grid's edge length $ e $, we first divide the space into combinations of grids, and then traverse every points in $ P $ to accumulate the number of points in each grid and compute its corresponding semantic label. In particular, assume that the vertex of the grid $ g $ with minimum coordinates is $ p_g = (x_g,y_g,z_g) $, the semantic of grid $ l_g $ can be defined as:}
\begin{align}
l_g &= \underset{l \in \mathcal{L}}{\arg\max} ( num(g,l) ) \notag \\
num(g,l) &= \left| p | p \in g, L(p)=l \right|,
\end{align}
{\color{zzx}where $ p \in g $ denotes the point $ p $ is inside the grid $ g $. If there exists two or more labels $\hat{\mathcal{L}} = \{ l_1, l_2, ... \}$ with the same number $ num(g,l) $, then the calculation of $ l_g $ will be updated as:}
\begin{equation}\label{key}
l_g = L(\underset{p}{\arg\min} ( dist(p,p_g) |  L(p) \in \hat{\mathcal{L}} )),
\end{equation}
{\color{zzx}where $ dist(p,p_g) $ denotes the European distance between point $ p $ and $ p_g $.}
	
{\color{zzx}For choosing an appropriate grid edge length $ e $, we calculate the $ CD $ between point cloud and its voxelization result as well as the voxelization time w.t.r six different $ e $ ($ 0.08,0.06,0.04,0.02,0.01,0.005 $ are used, equal to the resolution of $ 60\times 36 \times 60, 80 \times 48 \times 80,..., 960 \times 488 \times 960 $), as shown in Fig. \ref{fig:grid}. After balancing accuracy and efficiency, we finally exclude the case that $ e=0.005 $ and calculate the IoU for semantic segmentation on the other five $ e $. Following Song \textit{et al.} \cite{song2017semantic}, we do not evaluate on points outside the viewing frustum or the room.}

\noindent \textbf{Baselines}
This article is an extension of our previous CVPR work\cite{han2019oral}, thus one baseline is our previous work which is defined as Geometry Scene Completion(GSC). As shown in Tab. \ref{tab:pc_on_3df}, the input of GSC is only a single depth map, and the reinforcement learning network is DQN\cite{mnih2015human}. Also, we propose another baseline Colored Scene Completion(CSC) which introduce the RGB image into the algorithm and output a completed scene point cloud with plentiful color. The network compositions of GSC and CSC are similar to CSSC. In particular, GSC is composed of a volume completion network(SSCNet\cite{song2017semantic}), a depth inpainting network(PartialCNN\cite{liu2018image}), a projection layer connecting them and a MVCNN\cite{su15mvcnn} in DQN for determining the best geometry scene completion viewpoint path. The definition of DQN's component are almost same with A3C as mentioned in Sec. \ref{sec:a3c}, but only using $ R^{acc} $ and $ R^{hole} $ as its reward functions.The DQN takes mutil-view depth maps projected from $ P_{i-1} $ as inputs and outputs the Q-value of different actions. The whole network is trained to approximate the action-value function $ Q(P_{i-1},v_{i}) $, which is the expected reward that the agent receives when taking action $ v_{i} $ at state $ P_{i-1} $.
To ensure stability of the learning process, we introduce a target network separated from the architecture of \cite{mnih2015human}, whose loss function for training DQN is
\begin{equation}\label{equ5}
Loss(\theta)=\mathbb{E}[(r+\gamma \max\limits_{v_{i+1}}Q(P_{i},v_{i+1};\theta')-Q(P_{i-1},v_{i};\theta))^{2}].
\end{equation}
where $ r $ is the reward, $ \gamma $ a discount factor, and $ \theta' $ the parameters of the target network. For effective learning, we create an experience replay buffer to reduce the correlation between data. The buffer stores the tuples $(P_{i-1},v_{i},r,P_{i}) $ proceeded with the episode. We also employ the technique of \cite{van2016deep} to remove upward bias caused by $ \max_{v_{i+1}}Q(P_{i},v_{i+1};\theta') $ and change the loss function to
\begin{equation}\label{equ6}
\begin{split}
\mathbb{L}_{our}&=\mathbb{E}[(r+\gamma Q(P_{i},\arg\max_{v_{i+1}}Q(P_{i},v_{i+1};\theta);\theta')\\
&\quad-Q(P_{i-1},v_{i};\theta))^{2}].
\end{split}
\end{equation}

As for CSC baseline, the RGB information is introduced to the geometric completion method to predict the color of generated result. Compared with GSC, the differences are as follows: 1) We add an RGB image inpainting branch in depth inpainting module; 2) We update the reward function $ R^{acc} $ using in DQN by adding the $ L_1 $ loss between inpainted RGB image and its ground truth.

The only difference between CSSC baseline and our final approach is that: DQN is the algorithm used in View Path Planning Module of CSSC baseline while A3C is $ Ours $.

\begin{table*}[ht]	
	\centering
	\caption{Quantitative comparisons against existing methods and ablation studies on the 3D-FUTURE test set. The CD metric, {\color{zzx}$ C_r $ (Completeness) and $ A_r $ (Accuracy) (w.r.t different thresholds) are used. The units of $ C_r $ and $ A_r $ are percentages.}}
	\label{tab:pc_on_3df}

	\begin{tabular}{c|ccc|c|ccccc}
		\hline
		$ Methods $ & $ RGB $ & $ Seg. $ & $ RL $ & $CD$ & $ C_r / A_r(r=0.02) $ & $ C_r / A_r(0.04) $ & $ C_r / A_r(0.06) $ & $ C_r / A_r(0.08) $ & $ C_r / A_r(0.10) $ \\
		\hline
		\hline		
		$ SSCNet $ & $ \times $ & \checkmark & - & 2.2006 & 63.08/86.27 & 67.83/91.14 & 70.30/93.17 & 71.92/94.28 & 73.16/95.07\\
		$ VVNet $ & $ \times $ & \checkmark & - & 2.1832 & 66.45/\textbf{93.05} & 69.73/\textbf{95.02} & 71.59/\textbf{95.85} & 72.90/\textbf{96.35} & 73.95/\textbf{96.71} \\
		$ ForkNet $ & $ \times $ & \checkmark & - & 2.0818 & 67.09/91.05 & 70.46/93.92 & 72.27/95.12  & 73.60/95.88 & 74.56/96.38 \\
		$ GSC $  & $ \times $ & $ \times $ & $ DQN $ & 0.5364 & 77.25/73.55 & 80.82/79.00 & 82.86/81.96  & 84.29/83.92 & 85.39/85.37\\
		$ CSC $  & \checkmark & $ \times $ & $ DQN $ & 0.4503 & 77.62/77.04 & 81.72/78.61 & 84.06/82.38  & 85.66/84.93 & 86.88/86.78 \\
		$ CSSC $  & \checkmark & \checkmark & $ DQN $ & 0.4200 & 78.73/78.99 & 82.55/84.96 & 84.71/87.99 & 86.21/89.93 & 87.36/91.27\\
		$ Ours $  & \checkmark & \checkmark & $ A3C $ & \textbf{0.4095} & \textbf{79.78}/79.07 & \textbf{83.65}/85.03 & \textbf{85.87}/88.13 & \textbf{87.39}/90.11 & \textbf{88.55}/91.47\\
		\hline
		
		$ U_5 $ & \checkmark & \checkmark & - & 0.4520 & 78.61/79.31 & 82.39/85.15 & 84.49/88.22  & 85.99/90.15 & 87.14/91.49 \\
		$ U_{10} $ & \checkmark & \checkmark & - & 0.4476 & 79.23/77.48 & 82.89/83.79 & 84.99/87.06  & 86.47/89.12 & 87.59/90.55 \\
		$ U_{20} $ & \checkmark & \checkmark & - & 0.4576 & 78.54/77.90 & 82.41/83.88 & 84.62/87.07 & 86.14/89.10 & 87.30/90.54 \\
		$ Ours_{w/o.hole} $ & \checkmark & \checkmark & $ A3C $ & 0.4508 & 78.58/77.77 & 82.51/83.99 & 84.70/87.25 & 86.23/89.35 & 87.37/90.81\\
		$ Ours_{w/o.3D} $ & \checkmark & \checkmark & $ A3C $ & 0.4423 & 78.53/78.35 & 82.34/84.47 & 84.53/87.64 & 86.06/89.64 & 87.22/91.06\\
		\hline
	\end{tabular}
\end{table*}

\begin{table*}[htb]	
	\centering
	\caption{Quantitative semantic segmentation results in terms of {\color{zzx}scene completion $IoU$ ($ Com. $) and scene semantic $ IoU $ on the 3D-FUTURE test set}.}
	\label{tab:vox_on_3df}
	\begin{tabular}{|c|c|c|ccccccccccc|c|}
		\hline
		$ Methods $ & $ e $  & Com. & ceil. &  floor & wall & win. & chair & bed & sofa & table & tvs & furn. & objs. & $ Avg. $ \\
		\hline
		\hline		
		\multirow{5}{*}{SSCNet\cite{song2017semantic}}
		&$ e=0.01 $ & 1.17 & 0.08 & 0.16 & 1.35 & 0.27 & 0.22 & 0.07 & 0.18 & 0.13 & 0.46 & 0.59 & 0.51 & 0.36\\
		&$ e=0.02 $ & 2.44 & 0.22 & 0.27 & 2.68 & 0.55 & 0.44 & 0.11 & 0.41 & 0.31 & 0.63 & 1.34 & 1.00 & 0.72\\
		&$ e=0.04 $ & 4.22 & 0.70 & 0.80 & 4.12 & 0.95 & 0.81 & 0.21 & 0.71 & 0.72 & 0.84 & 2.07 & 1.83 & 1.25\\
		&$ e=0.06 $ & 8.51 & 5.06 & 4.30 & 7.39 & 2.13 & 1.39 & 0.36 & 1.18 & 1.56 & 1.51 & 3.36 & 3.42 & 2.88\\
		&$ e=0.08 $ & 14.85 & 0.99 & 14.21 & 6.29 & 2.28 & 2.34 & 1.19 & 2.60 & 3.87 & 1.27 & 5.55 & 4.80 & 4.13\\
		\hline
		
		\multirow{5}{*}{VVNet\cite{guo2018view}}	
		&$ e=0.01 $ & 1.33 & 0.04 & 0.19 & 1.76 & 0.37 & 0.23 & 0.09 & 0.21 & 0.19 & 0.32 & 0.47 & 0.68 & 0.41\\
		&$ e=0.02 $ & 2.90 & 0.10 & 0.32 & 3.39 & 0.91 & 0.53 & 0.15 & 0.53 & 0.46 & 0.51 & 1.31 & 1.51 & 0.88\\
		&$ e=0.04 $ & 5.26 & 0.31 & 0.92 & 5.60 & 1.57 & 1.04 & 0.33 & 0.97 & 1.32 & 0.78 & 2.19 & 2.69 & 1.61\\
		&$ e=0.06 $ & 10.82 & 3.17 & 6.41 & 10.47 & 3.63 & 1.70 & 0.62 & 1.85 & 2.25 & 1.65 & 3.81 & 5.43 & 3.73\\
		&$ e=0.08 $ & 18.68 & 0.48 & 21.73 & 8.85 & 3.44 & 2.84 & 1.79 & 3.39 & 4.89 & 1.29 & 6.59 & 7.49 & 5.71\\		
		\hline
		
		\multirow{5}{*}{ForkNet\cite{wang2019forknet}}	
		&$ e=0.01 $ & 2.32 & \textbf{2.77} & 1.70 & 2.45 & 0.28 & 0.51 & 0.23 & 0.39 & 1.26 & 0.26 & 1.11 & 1.39 & 1.12\\
		&$ e=0.02 $ & 4.98 & \textbf{4.92} & 2.87 & 5.08 & 0.63 & 1.31 & 0.53 & 1.07 & 1.87 & 0.53 & 2.76 & 2.94 & 2.23\\
		&$ e=0.04 $ & 16.68 & 6.05 & \textbf{45.31} & 10.77 & 1.57 & 2.54 & 1.51 & 1.96 & 3.62 & 1.73 & 5.13 & 6.20 & 7.85\\
		&$ e=0.06 $ & 20.63 & 5.75 & \textbf{47.14} & 10.50 & 2.14 & 3.75 & 2.36 & 3.97 & 11.81 & 2.03 & 9.23 & 8.35 & 9.73\\
		&$ e=0.08 $ & 27.87 & 6.93 & \textbf{44.44} & 20.91 & 4.89 & 4.80 & 2.68 & 5.01 & 7.67 & 6.05 & 11.16 & 11.70 & 11.47\\		
		\hline
		
		\multirow{5}{*}{$ CSSC $}	
		&$ e=0.01 $ & 20.99 & 1.42 & 16.80 & \textbf{18.35} & 12.23 & 7.21 & \textbf{4.17} & \textbf{5.70} & \textbf{9.48} & \textbf{8.87} & 14.04 & 18.54 & 10.62\\
		&$ e=0.02 $ & 25.49 & 3.40 & 20.71 & 21.10 & 14.63 & 7.84 & 5.18 & 7.24 & 10.24 & \textbf{11.17} & 17.54 & 21.56 & 12.78\\
		&$ e=0.04 $ & 29.65 & 5.90 & 25.11 & 22.45 & 15.29 & 8.04 & \textbf{6.21} & 8.00 & 11.02 & \textbf{11.26} & 19.82 & 23.22 & 14.21\\
		&$ e=0.06 $ & 32.25 & 6.16 & 26.43 & 23.71 & 16.24 & 8.20 & 6.23 & 8.53 & 11.28 & 11.42 & 20.94 & 23.57 & 14.79\\
		&$ e=0.08 $ & 34.89 & 6.59 & 29.10 & 25.06 & 16.08 & 8.23 & 6.54 & 8.94 & 11.85 & 10.69 & 21.38 & 24.24 & 15.34\\		
		\hline
		
		\multirow{5}{*}{$ Ours $}	
		&$ e=0.01 $ & \textbf{21.07} & 1.43 & \textbf{16.88} & 18.34 & \textbf{12.30} & \textbf{7.25} & 4.12 & 5.66 & 9.34 & 8.79 & \textbf{14.14} & \textbf{18.92} & \textbf{10.65}\\
		&$ e=0.02 $ & \textbf{26.02} & 3.55 & \textbf{20.91} & \textbf{21.50} & \textbf{15.18} & \textbf{8.03} & \textbf{5.21} & \textbf{7.34} & \textbf{10.33} & 11.11 & \textbf{17.66} & \textbf{22.26} & \textbf{13.01}\\
		&$ e=0.04 $ & \textbf{30.85} & \textbf{6.26} & 25.78 & \textbf{23.23} & \textbf{16.45} & \textbf{8.50} & 5.98 & \textbf{8.26} & \textbf{11.51} & 11.23 & \textbf{20.26} & \textbf{24.56} & \textbf{14.73}\\
		&$ e=0.06 $ & \textbf{33.89} & \textbf{6.65} & 27.26 & \textbf{24.65} & \textbf{18.20} & \textbf{8.79} & \textbf{6.46} & \textbf{8.93} & \textbf{11.91} & \textbf{11.72} & \textbf{21.97} & \textbf{25.44} & \textbf{15.63}\\
		&$ e=0.08 $ & \textbf{36.96} & \textbf{7.27} & 30.37 & \textbf{26.16} & \textbf{17.71} & \textbf{8.96} & \textbf{6.86} & \textbf{9.58} & \textbf{12.61} & \textbf{11.37} & \textbf{22.26} & \textbf{26.47} & \textbf{16.33}\\		
		\hline	
		
	\end{tabular}
\end{table*}

\subsection{Quantitative Comparisons}
We validate the effectiveness of the proposed approach by comparing the state-of-the-art methods, including SSCNet \cite{song2017semantic}, VVNet \cite{guo2018view} and ForkNet \cite{wang2019forknet}. {\color{zzx}For fair evaluation, we first retrained the above 3 methods on the 3D-FUTURE dataset using their respective hyperparameter settings mentioned in \cite{song2017semantic, guo2018view, wang2019forknet}. In particular, since the generated training dataset of 3D-FUTURE only contains 1,431 images, training these networks on this dataset alone is insufficient. As a solution already used in \cite{wang2019forknet}, we initialize parameters of these 3 networks by the provided pre-trained model, and then refine them by supplementing the training data from 3D-FUTURE with 1,400 randomly selected samples from their original datasets in each epoch of training.} After finishing training process, we first render the volume obtained from SSCNet, VVNet and ForkNet to several depth maps under the same viewpoints as our method. We then convert these depth maps to point cloud. 

The results on the 3D-FUTURE dataset are shown in Tab. \ref{tab:pc_on_3df}. As seen, our approach outperforms all the others on $ CD $ and $ Completeness $ metrics. {\color{zzx}Since the $ Accuracy $ metric measures the accuracy of the prediction points, a high $ A_r $ with low $ C_r $ indicates that the algorithm tends to output only the points with high confidence and ignore the completeness of the model.} This also validates that the using of volumetric representation greatly reduces the quality of the outputs. Considering that it may be unfair to convert voxel predictions to point cloud, we convert our point cloud result to voxels by judging whether there is a point within the grid. However, compared to voxel results output from SSCNet, VVNet and ForkNet, our predicted point cloud only provides surface information. It means converting a point cloud to voxel and directly comparing to the corresponding voxel ground truth is also unfair. Therefore, to enforce fairness, we transform our reconstructed point cloud to voxel and compare them with the corresponding ground truth that is on the surface only.

 These voxel-based results are reported in Tab. \ref{tab:vox_on_3df} evaluated by each class and average IoU for semantic segmentation task.{ \color{zzx}Following Song \textit{et al.} \cite{song2017semantic}, we also report the scene completion performance of related methods by treating all non-empty object class as one category and evaluating IoU of the binary predictions. Except classes $ ceil $ and $ floor $, our approach provides more accurate results compared with voxel-based methods. It also can be observed that the IoU of voxel-based methods increase significantly when $ e $ increases from $ 0.02 $ to $ 0.04 $ or from $ 0.04 $ to $ 0.06 $, which indicates that the predictions of such methods are far from the ground truth surface of the scene and sufficient accuracy can only be obtained at the expense of high resolution. In contrast, our performance grows more smoothly, indicating that our generated points are closer to the ground truth scene surface.}

We also report different tasks' completion results in Tab \ref{tab:pc_on_3df}: Geometric Scene Completion(GSC) which only completes the 3D points, Colored Scene Completion(CSC) which will also output the completed points' colored labels, Colored Semantic Scene Completion(CSSC) which outputs both the colored label and semantic label of the point but uses DQN as its view planning method and $ Ours $ which output the same as CSSC with A3C as the RL method. And we can draw a conclusion: with the input of more information and the output of multi-tasking and the adjustment of the algorithm structure, the performance of scene completion is continuously improving.

{\color{zzx}The comparison of results on the ScanNet dataset are shown in Tab. \ref{tab:pc_on_scan} and Tab. \ref{tab:vox_on_scan}. The first observation is that the performance of all methods on the real data degrades, especially for SSCNet and VVNet. It is more clearly observed in the visualization results in Fig. \ref{fig:sota_real} that these two methods are no longer available. This also proves that ForkNet and our method have the better generalization property. Since the ground truth models in the ScanNet dataset do not contain ceilings, the IoU of class $ ceil. $ in Tab. \ref{tab:vox_on_scan} are all zeros.}

\begin{table*}[htb]	
	\centering
	\caption{{\color{zzx}Quantitative comparisons against existing methods and ablation studies on the ScanNet test set.} The CD metric, $ C_r $ (Completeness) and $ A_r $ (Accuracy) (w.r.t different thresholds) are used. The units of $ C_r $ and $ A_r $ are percentages.}
	\label{tab:pc_on_scan}

	\begin{tabular}{c|ccc|c|ccccc}
		\hline
		$ Methods $ & $ RGB $ & $ Seg. $ & $ RL $ & $CD$ & $ C_r / A_r(r=0.02) $ & $ C_r / A_r(0.04) $ & $ C_r / A_r(0.06) $ & $ C_r / A_r(0.08) $ & $ C_r / A_r(0.10) $ \\
		\hline
		\hline		
		$ SSCNet $ & $ \times $ & \checkmark & - & 5.0607 & 40.17/22.97 & 50.76/30.70 & 57.77/35.47 & 62.85/38.87 & 66.67/41.50\\
		$ VVNet $ & $ \times $ & \checkmark & - & 4.8375 & 40.07/56.12 & 48.27/66.87 & 53.89/71.61 & 58.31/74.22 & 61.87/76.03 \\
		$ ForkNet $ & $ \times $ & \checkmark & - & 2.1234 & 64.96/\textbf{67.95} & 69.74/\textbf{71.67} & 72.08/\textbf{73.24}  & 73.58/\textbf{74.32} & 74.52/\textbf{75.01} \\
		$ Ours $  & \checkmark & \checkmark & $ A3C $ & \textbf{0.8635} & \textbf{69.02}/62.88 & \textbf{71.91}/66.59 & \textbf{73.22}/68.62 & \textbf{74.08}/69.94 & \textbf{74.66}/70.85\\
		\hline
	\end{tabular}
\end{table*}

\begin{table*}[htb]	
	\centering
	\caption{Quantitative semantic segmentation results in terms of scene completion $IoU$ ($ Com. $) and scene semantic $ IoU $ {\color{zzx}on the ScanNet test set.}}
	\label{tab:vox_on_scan}
	\begin{tabular}{|c|c|c|ccccccccccc|c|}
		\hline
		$ Methods $ & $ e $  & Com. & ceil. &  floor & wall & win. & chair & bed & sofa & table & tvs & furn. & objs. & $ Avg. $ \\
		\hline
		\hline		
		\multirow{5}{*}{SSCNet\cite{song2017semantic}}	
		&$ e=0.01 $ & 0.72 & 0.00 & 0.43 & 0.01 & 0.00 & 0.00 & 0.00 & 0.02 & 0.02 & 0.00 & 0.07 & 0.10 & 0.06\\
		&$ e=0.02 $ & 1.33 & 0.00 & 0.92 & 0.01 & 0.00 & 0.00 & 0.00 & 0.05 & 0.04 & 0.00 & 0.13 & 0.17 & 0.12\\
		&$ e=0.04 $ & 2.42 & 0.00 & 1.90 & 0.03 & 0.00 & 0.00 & 0.00 & 0.05 & 0.10 & 0.00 & 0.27 & 0.26 & 0.24\\
		&$ e=0.06 $ & 3.24 & 0.00 & 2.47 & 0.05 & 0.00 & 0.00 & 0.01 & 0.06 & 0.16 & 0.00 & 0.40 & 0.25 & 0.31\\
		&$ e=0.08 $ & 4.35 & 0.00 & 4.28 & 0.09 & 0.00 & 0.00 & 0.00 & 0.04 & 0.26 & 0.00 & 0.62 & 0.39 & 0.52\\		
		\hline
		
		\multirow{5}{*}{VVNet\cite{guo2018view}}
		&$ e=0.01 $ & 1.41 & 0.00 & 1.17 & 0.01 & 0.01 & 0.12 & 0.00 & 0.01 & 0.15 & 0.00 & 0.00 & 0.18 & 0.15\\
		&$ e=0.02 $ & 2.40 & 0.00 & 2.27 & 0.02 & 0.00 & 0.19 & 0.00 & 0.01 & 0.22 & 0.00 & 0.00 & 0.33 & 0.28\\
		&$ e=0.04 $ & 4.19 & 0.00 & 4.43 & 0.05 & 0.00 & 0.31 & 0.00 & 0.02 & 0.32 & 0.00 & 0.00 & 0.56 & 0.52\\
		&$ e=0.06 $ & 5.47 & 0.00 & 5.85 & 0.03 & 0.00 & 0.40 & 0.00 & 0.01 & 0.35 & 0.00 & 0.00 & 0.87 & 0.68\\
		&$ e=0.08 $ & 7.81 & 0.00 & 11.58 & 0.07 & 0.00 & 0.59 & 0.00 & 0.04 & 0.66 & 0.00 & 0.00 & 0.81 & 1.25\\		
		\hline
		
		\multirow{5}{*}{ForkNet\cite{wang2019forknet}}	
		&$ e=0.01 $ & 3.88 & 0.00 & 0.99 & 1.82 & 0.00 & 0.35 & 0.04 & 0.64 & 0.13 & \textbf{0.01} & 0.62 & 0.73 & 0.49\\
		&$ e=0.02 $ & 7.13 & 0.00 & 1.94 & 2.72 & 0.01 & 0.65 & 0.08 & 1.18 & 0.27 & 0.00 & 1.19 & 1.35 & 0.85\\
		&$ e=0.04 $ & 12.11 & 0.00 & 3.24 & 4.24 & 0.04 & 1.20 & 0.13 & 1.94 & 0.58 & 0.00 & 1.88 & 2.13 & 1.40\\
		&$ e=0.06 $ & 16.93 & 0.00 & 5.79 & 4.68 & 0.04 & 1.79 & 0.28 & 2.48 & 1.09 & 0.19 & 2.48 & 2.77 &1.96 \\
		&$ e=0.08 $ & 20.09 & 0.00 & 5.33 & 5.32 & 0.00 & 1.62 & 0.54 & 3.00 & 1.50 & 0.00 & 2.65 & 3.24 & 2.11\\		
		\hline
		
		\multirow{5}{*}{$ Ours $}	
		&$ e=0.01 $ & \textbf{7.73} & 0.00 & \textbf{7.95} & \textbf{4.90} & \textbf{0.77} & \textbf{3.05} & \textbf{0.84} & \textbf{5.77} & \textbf{5.12} & 0.00 & \textbf{4.36} & \textbf{2.56} & \textbf{3.21}\\
		&$ e=0.02 $ & \textbf{12.74} & 0.00 & \textbf{14.31} & \textbf{9.01} & \textbf{1.16} & \textbf{4.99} & \textbf{1.68} & \textbf{9.32} & \textbf{8.38} & 0.00 & \textbf{8.05} & \textbf{4.19} & \textbf{5.55}\\
		&$ e=0.04 $ & \textbf{17.39} & 0.00 & \textbf{19.60} & \textbf{12.79} & \textbf{1.85} & \textbf{5.84} & \textbf{3.02} & \textbf{11.74} & \textbf{10.58} & \textbf{0.17} & \textbf{13.08} & \textbf{5.40} & \textbf{7.64}\\
		&$ e=0.06 $ & \textbf{20.22} & 0.00 & \textbf{22.05} & \textbf{14.69} & \textbf{1.54} & \textbf{6.76} & \textbf{2.94} & \textbf{12.39} & \textbf{12.11} & \textbf{0.38} & \textbf{15.95} & \textbf{6.28} & \textbf{8.64}\\
		&$ e=0.08 $ & \textbf{22.43} & 0.00 & \textbf{24.07} & \textbf{16.31} & \textbf{1.64} & \textbf{6.43} & \textbf{3.92} & \textbf{13.97} & \textbf{13.09} & \textbf{0.74} & \textbf{17.17} & \textbf{6.73} & \textbf{9.46}\\		
		\hline
	\end{tabular}
\end{table*}

\subsection{Qualitative Comparisons}
\begin{figure}[htb]
	\centering
	\includegraphics[width=0.475\textwidth]{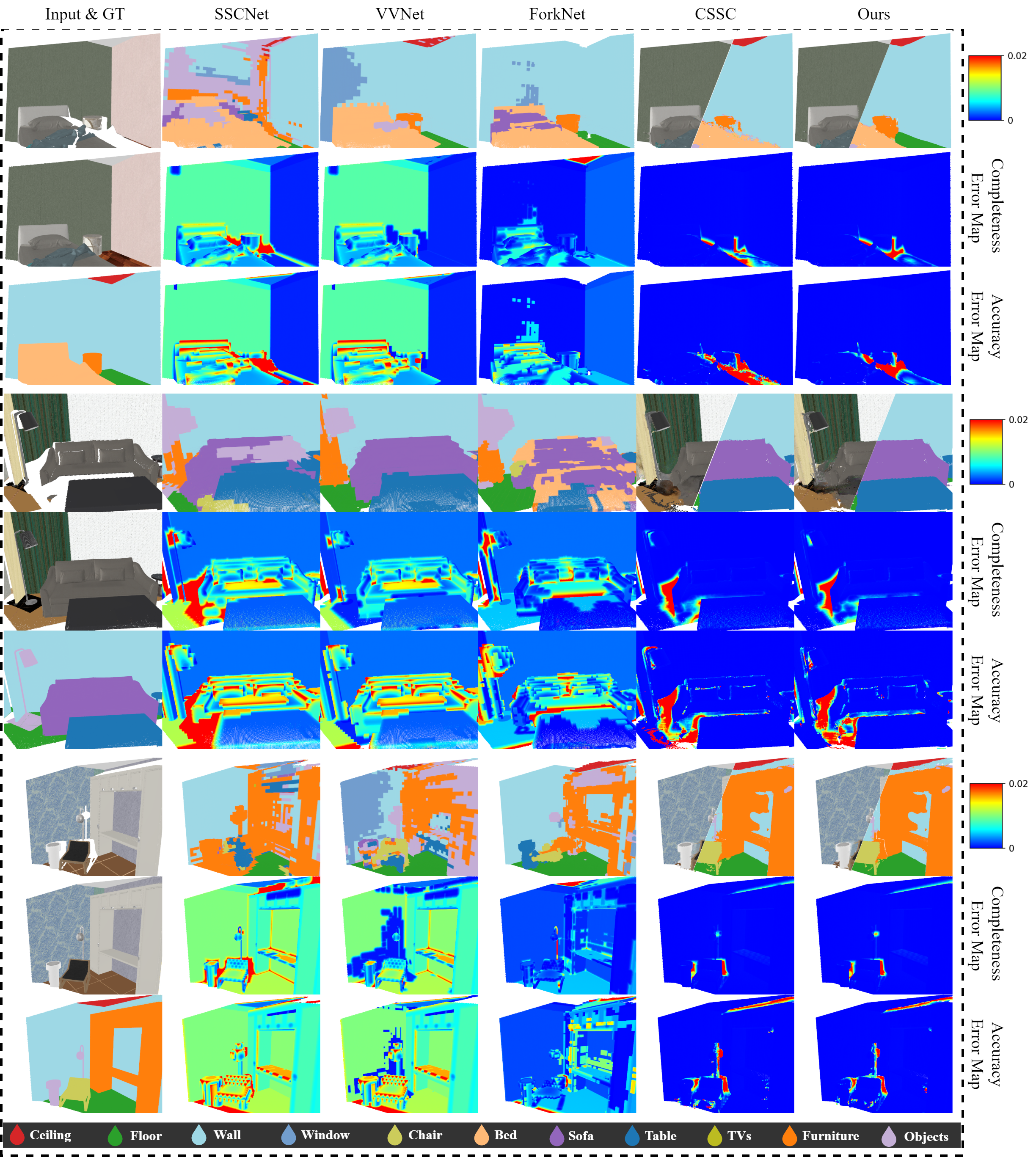}
	\caption{Comparisons against the state-of-the-arts on the 3D-FUTURE test set. Given different inputs and the referenced groundtruth, we show the semantic completion results of four modules, {\color{zzx}with the corresponding point cloud completeness and accuracy error maps below.}}
	\label{fig:sotapc}
	\vspace{-0.5cm}
\end{figure}

\begin{figure}[ht]
	\centering
	\includegraphics[width=0.475\textwidth]{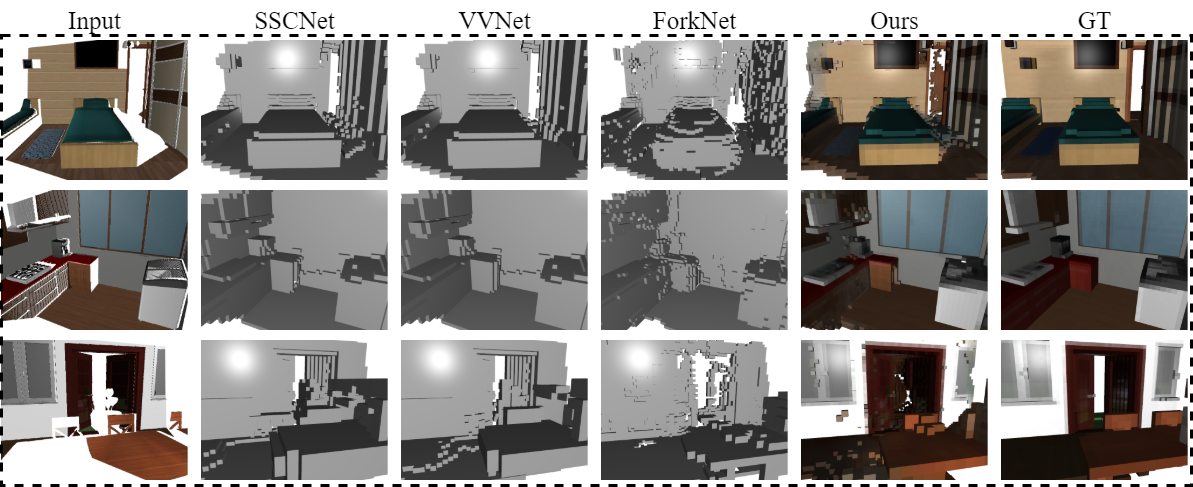}
	\caption{Comparisons against the state-of-the-arts with voxel representation.}
	\label{fig:sotavox}
	\vspace{-0.5cm}
\end{figure}

\begin{figure}[ht]
	\centering
	\includegraphics[width=0.475\textwidth]{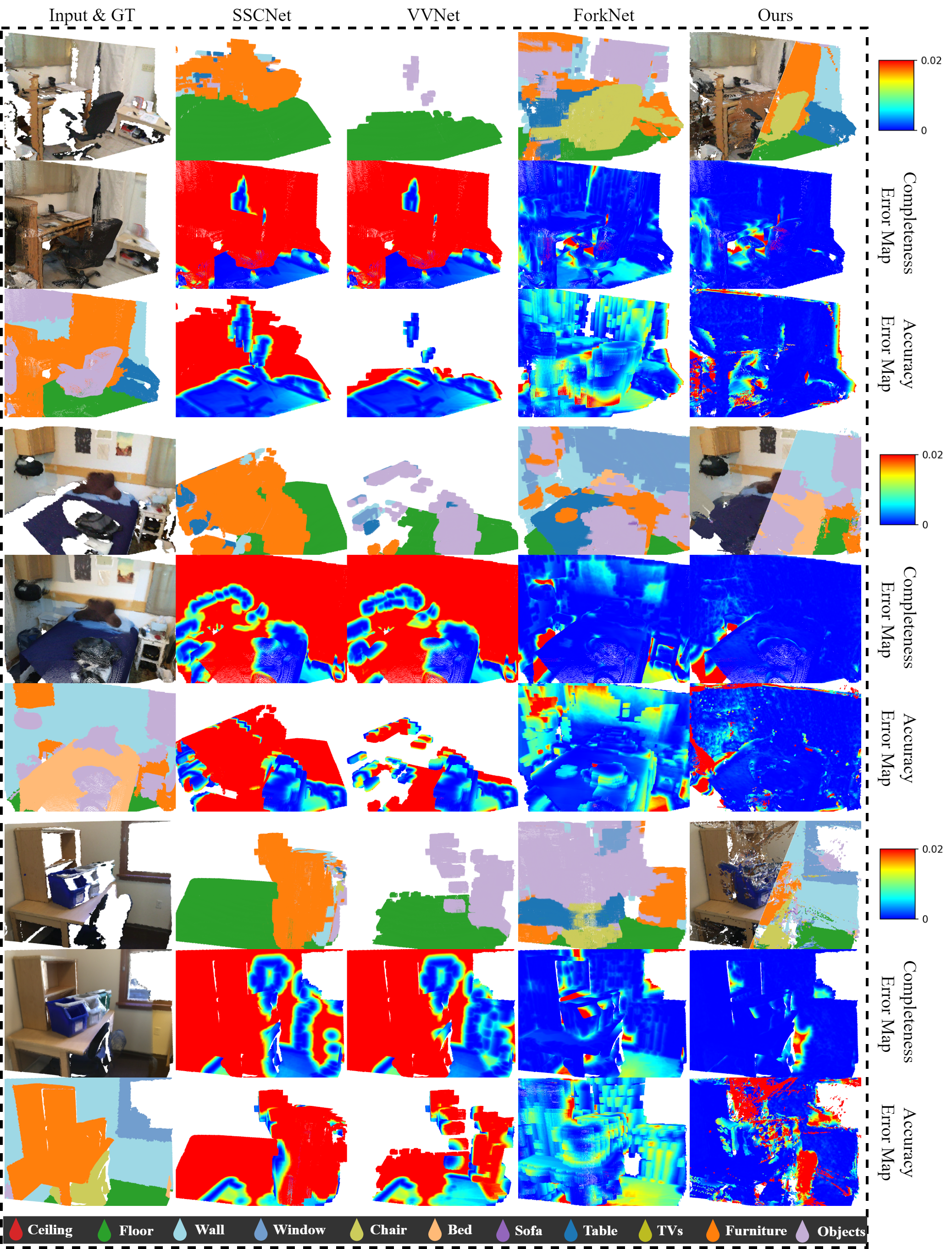}
	\caption{{\color{zzx}Scene semantic completion results against the state-of-the-arts on the ScanNet test set, with the corresponding point cloud completeness and accuracy error maps below.}}
	\label{fig:sota_real}
	\vspace{-0.5cm}
\end{figure}

The visual comparisons of these methods are shown in Fig. \ref{fig:sotapc} and Fig. \ref{fig:sota_real}. It can be seen that, the generated point clouds from voxel-based methods are of no surface details. Our proposed method not only produces more accurate results, but also predicts the reasonable colors for generating points. This can be validated in Tab \ref{tab:pc_on_3df} and shown in Fig \ref{fig:sotapc}. In addition, by conducting completion in multiple views, our approach also recovers more missing points, showing better completeness as validated in Tab \ref{tab:pc_on_3df}. We show the colored labeled completion results of CSC in Fig. \ref{fig:ablation}. We can observe how additional information and view planing method influence the completion performance from the perspective of visualization. We also show our volumetric results which are converting from point cloud in Fig. \ref{fig:sotavox}.

\subsection{Ablation Studies}
\label{sec:ablation}

To ensure the effectiveness of several key components of our system, we do some control experiments by removing each component.

\begin{table*}[htb]
	\centering
	\caption{Quantitative ablation studies on RGB-D image inpainting network.}
	\label{tab:ab_inp}
	\begin{tabular}{|c|c|cccccc|ccc|}
		\hline
		$ Task $ & $ Version $ & $ Backbone $ & $ Seg. $ & $ Depth $ & $ RGB $ & $ V_{dep} $  & $ PBP $ & $L^1_\Omega$ & $ PSNR $ & $ SSIM $  \\
		\hline
		\hline
		\multirow{6}{*}{\shortstack{Depth \\ Inpainting }} & $ Case1 $ & $ PartConv $ & $ \times $ & - & $ \times $ &$ \times $ & $ \times $ & 0.0805 & 25.52 & 0.9533\\
		& $ Case2 $ & $ PartConv $ & \checkmark & - & $ \times $ & $ \times $  & $ \times $ & 0.0653 & 27.41 & 0.9512\\
		& $ Case3 $ & $ StrucFlow $ & \checkmark & - & $ \times $ & $ \times $  & $ \times $ & 0.0631 & 26.89 & 0.9443\\
		& $ Case4 $ & $ PartConv $ & \checkmark & - & \checkmark & $ \times $  & $ \times $ & 0.0615 & 27.31 & 0.9528\\
		& $ Case5 $ & $ PartConv $ & \checkmark & - & \checkmark & \checkmark  & $ \times $ & 0.0593 & 27.98 & 0.9582\\
		& $ Case6 $ & $ PartConv $ & \checkmark & - & \checkmark & \checkmark  & \checkmark & \textbf{0.0543} & \textbf{28.56} & \textbf{0.9608}\\
		
		\hline
		\multirow{5}{*}{\shortstack{RGB \\ Inpainting }} & $ Case7 $ & $ PartConv $ & $ \times $ & $ \times $ & - & -  & $ \times $ & 0.1056 & 23.93 & 0.9064\\
		& $ Case8 $ & $ PartConv $ & \checkmark & $ \times $ & - & - & $ \times $ & 0.0986 & 24.44 & 0.9124\\
		& $ Case9 $ & $ StrucFlow $ & \checkmark & $ \times $ & - & - & $ \times $ & 0.0683 & 25.77 & 0.8971\\
		& $ Case10 $ & $ StrucFlow $ & \checkmark & \checkmark & - & -  & $ \times $ & 0.0603 & 26.38 & 0.8947\\
		& $ Case11 $ & $ StrucFlow $ & \checkmark & \checkmark & - & -  & \checkmark & \textbf{0.0587} & \textbf{27.03} & \textbf{0.9288}\\
		\hline
		
	\end{tabular}
\end{table*}

\subsubsection{On Image Inpainting}
\label{abs:img}

\begin{figure}[ht]
	\centering
	\includegraphics[width=0.495\textwidth]{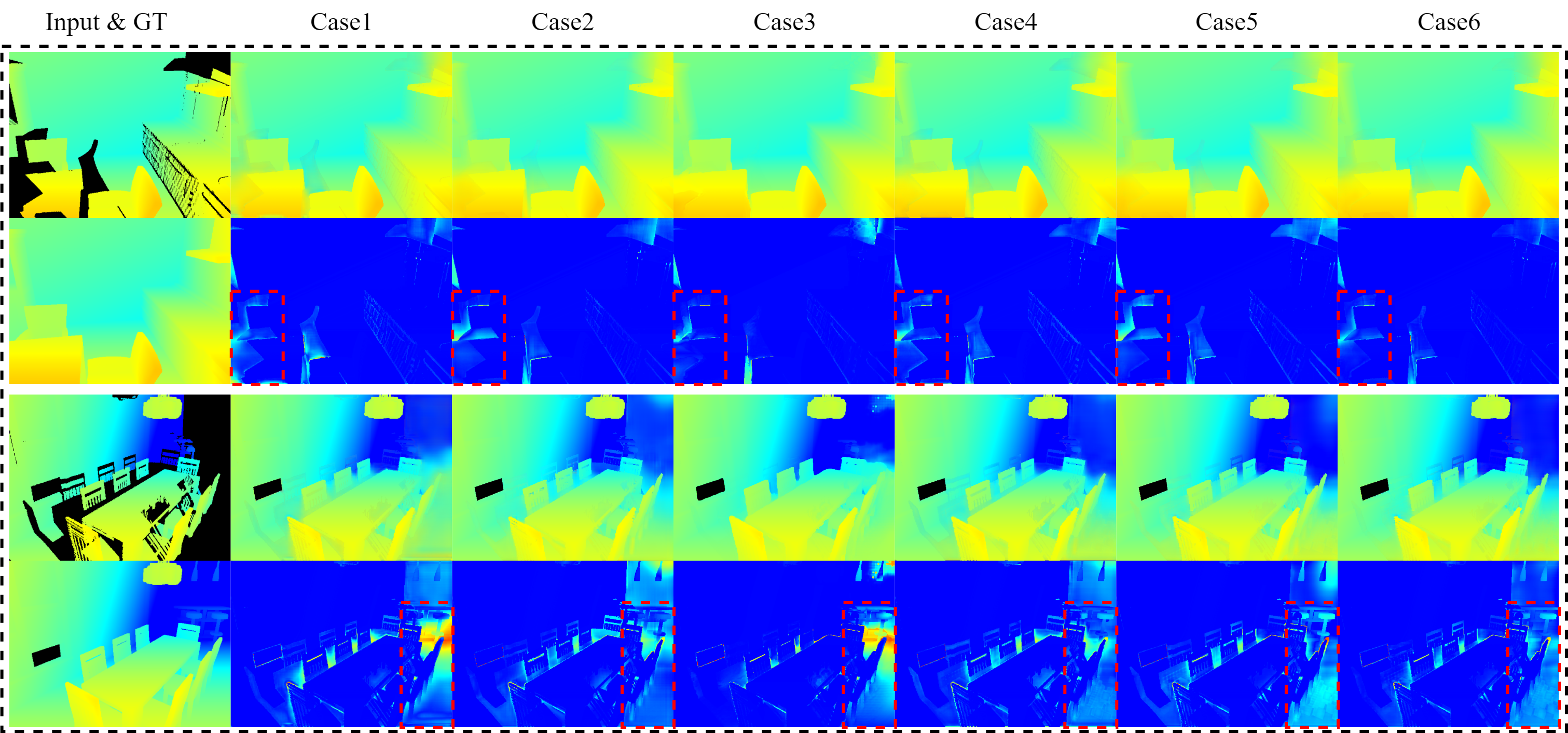}
	\caption{Comparisons on the variants of depth inpainting network. Given incomplete depth images, we show results of different case compared with the groundtruth. Both the inpainted map and its error map are shown.}
	\label{fig:2dab_dep}
	\vspace{-0.5cm}
\end{figure}

\begin{figure}[ht]
	\centering
	\includegraphics[width=0.475\textwidth]{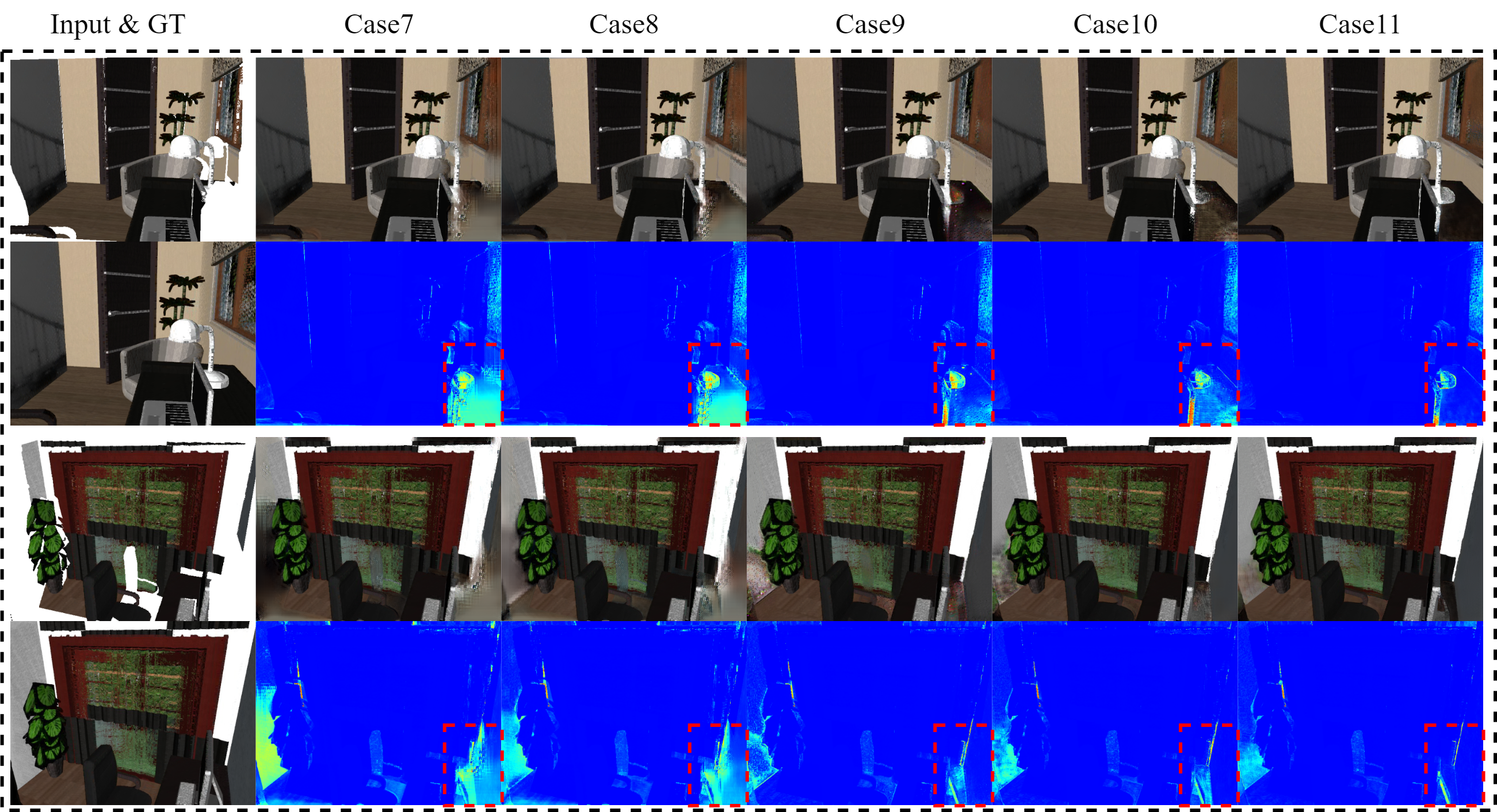}
	\caption{Comparisons on the variants of RGB inpainting network. Given incomplete RGB images, we show results of different case compared with the groundtruth. Both the inpainted map and its error map are shown.}
	\label{fig:2dab_rgb}
	\vspace{-0.5cm}
\end{figure}

\begin{figure}[ht]
	\centering
	\includegraphics[width=0.475\textwidth]{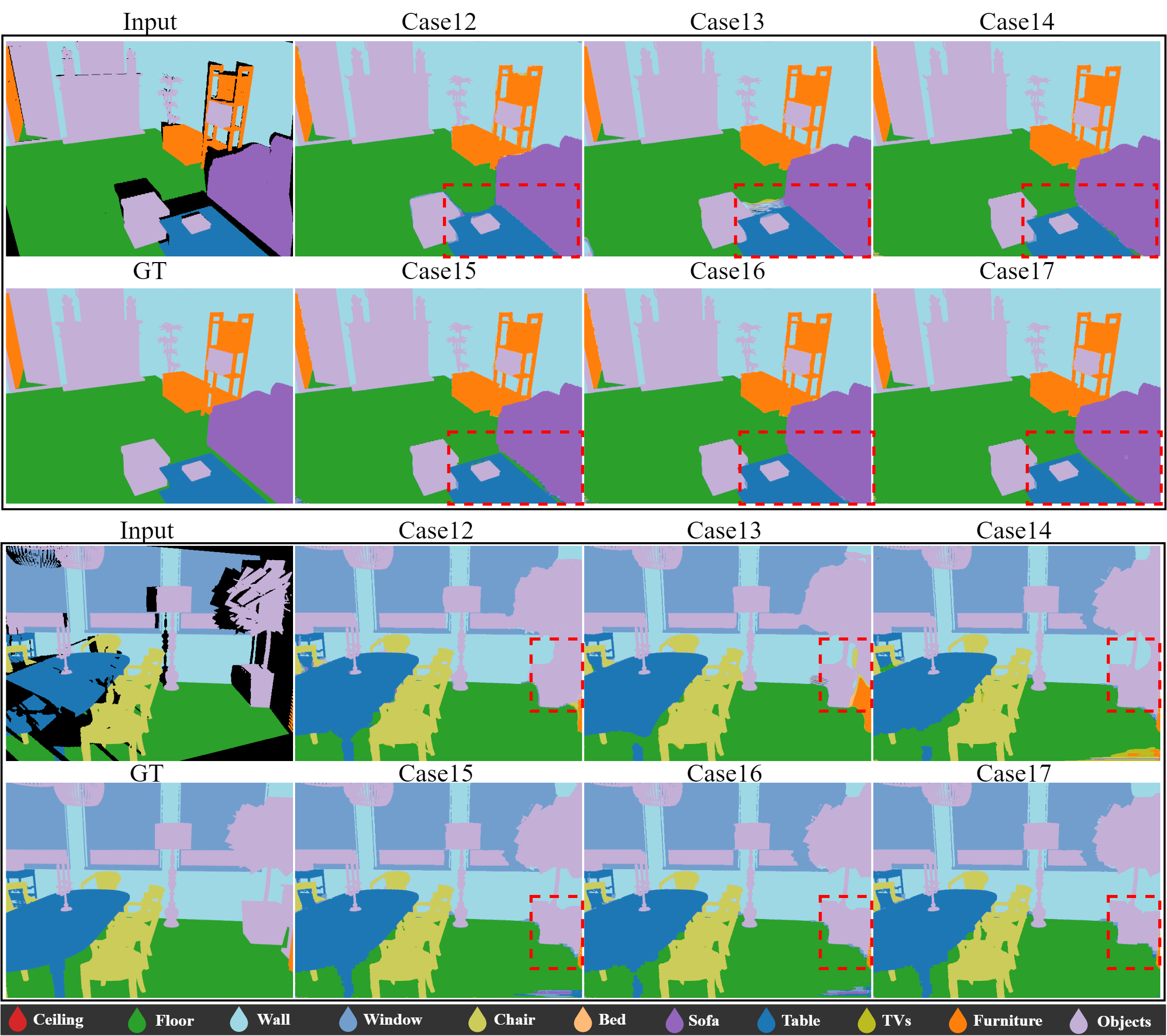}
	\caption{Comparisons on the variants of segmentation inpainting network. Given incomplete segmentation images, we show results of different case compared with the groundtruth.}
	\label{fig:2dab_seg}
	\vspace{-0.5cm}
\end{figure}

We ablate our 2D image inpainting method with different configurations as shown in Tab. \ref{tab:ab_inp} ,Tab. \ref{tab:seg_inp}, Fig. \ref{fig:2dab_dep}, Fig. \ref{fig:2dab_rgb} and Fig. \ref{fig:2dab_seg}. The $ BN $ in Tab. \ref{tab:seg_inp} denotes $ Backbone $, and $ P.C. $ for $ PartConv $, $ S.F. $ for $ StrucFlow $. {\color{zzx}The $ PBP $ in Tab. \ref{tab:ab_inp} and Tab. \ref{tab:seg_inp} denotes whether the volume-guided 2D inpainting network is trained with or without projection back-propagation.}

We use the metrics of $ L_{\Omega}^{1} $, $ PSNR $ and $ SSIM $ for the RGB-D inpainting comparisons, $ IoUs $ for semantic inpainting comparisons. The quantitative results are reported in Tab. \ref{tab:ab_inp} and Tab. \ref{tab:seg_inp}, from which we observe that: \\
\bm{$ Case2 $} v.s. \bm{$ Case3 $}, \bm{$ Case8 $} v.s. \bm{$ Case9 $}: A key idea in StrucFlow is to use appearance flow to sample features from regions with similar structures \cite{ren2019structureflow}, which means that the network could easily predict the color detail if it can refer to areas with the same structure. This can explain the following situation: The RGB inpainting network using StructureFlow baseline is better than the one using PartialCNN baseline, but at the same time the depth inpainting network using PartialCNN is better then StructureFlow. Areas with similar structures in RGB images generally have similar colors, but areas with similar structures in depth images may indeed have large differences in depth values.

\noindent From \bm{$ Case1 $} to \bm{$ Case5 $}: With the increase of effective input(incomplete RGB image, predicted segmentation image and depth map rendered from voxel completion result), the effect of depth map completion gradually improves. Especially in scene one (Fig. \ref{fig:2dab_dep} top 2rows), the network can gradually predict the information behind the hole in the back of the chair(inside the red dashed box).

\noindent From \bm{$ Case7 $} to \bm{$ Case10 $}: With the increase of effective input(incomplete depth image and predicted segmentation image), the effect of RGB image completion gradually improves.

\noindent From \bm{$ Case12 $} to \bm{$ Case16 $}: With the increase of effective input(incomplete depth image, incomplete RGB image and and segmentation map rendered from voxel completion result), the effect of segmentation image completion gradually improves.

\noindent \bm{$ Case6 $}, \bm{$ Case11 $} and \bm{$ Case17 $}: The projection layer used for joint training do help increase the accurate of RGB-D image inpainting. But the improvement of segmentation inpainting is not obvious, this may be because the semantic segmentation map contains a lot of structure information, and it is not suitable to use the method of inpainting ordinary images to complete it. Combine with the previous observation, the efficacy of the volume guidance is evaluated.

\begin{table*}[htb]
	\centering
	\caption{Quantitative ablation studies on segmentation image inpainting network.}
	\label{tab:seg_inp}
	\begin{tabular}{|c|ccccc|cccccccccccc|}
		\hline
		$ Version $ & $ BN $ & $ Dep. $ & $ RGB $ & $ V_{seg} $ & $ PBP $ & ceil. &  floor & wall & win. & chair & bed & sofa & table & tvs & furn. & objs. & $ Avg. $  \\
		\hline
		\hline
		
		$ Case12 $ & $ P.C. $ & $ \times $ & $ \times $ & $ \times $ & $ \times $ & 7.63 & 56.28 & 79.44 & 50.28 & 25.57 & 33.39 & 21.56 & 41.28 & 26.67 & 51.80 & 74.44 & 42.58\\
		$ Case13 $ & $ S.F. $ & $ \times $ & $ \times $ & $ \times $ & $ \times $ & 7.55 & 59.23 & 75.38 & 50.47 & \textbf{27.88} & 30.31 & 20.11 & 39.57 & 32.14 & 52.09 & 69.72 & 42.22\\
		$ Case14 $ & $ P.C. $ & \checkmark & $ \times $ & $ \times $ & $ \times $ & 9.08 & 58.77 & 80.16 & 48.41 & 25.88 & 32.13 & 22.16 & 42.42 & 29.17 & 51.28 & 75.66 & 43.19\\
		$ Case15 $ & $ P.C. $ & \checkmark & \checkmark & $ \times $ & $ \times $ & 9.09 & 59.36 & 80.57 & 47.89 & 26.03 & 32.11 & 21.93 & 42.55 & 28.51 & 51.42 & 76.01 & 43.22\\
		$ Case16 $ & $ P.C. $ & \checkmark & \checkmark & \checkmark & $ \times $ & 9.08 & \textbf{60.31} & \textbf{81.23} & \textbf{54.93} & 24.55 & 31.28 & \textbf{23.84} & \textbf{43.76} & 21.30 & 50.48 & \textbf{77.41} &  43.47\\
		$ Case17 $ & $ P.C. $ & \checkmark & \checkmark & \checkmark & \checkmark & \textbf{9.21} & 58.77 & 79.44 & 51.18 & 26.31 & \textbf{34.40} & 22.31 & 42.59 & \textbf{29.72} & \textbf{52.40} & 76.50 & \textbf{43.89}\\
		\hline
		
	\end{tabular}
\end{table*}

\subsubsection{On Action Space Setting}

{\color{zzx}In order to determine the effect of different $ (\theta, \phi) $ on the completion performance, we additionally set 3 new sets of action spaces and retrain the A3C algorithm, and the final reconstruction results are shown in Tab. \ref{tab:actionspace}. As can be seen, $ \theta $ has little effect on the reconstruction performance ( $ Ours $ v.s. $ Case a $ and $ Case b $ v.s. $ Case c $), while a too small $ \phi $ will lead to a worse situation ( $ Ours $ v.s. $ Case b $ and $ Case a $ v.s. $ Case c $).
}

\subsubsection{On View Path Planning}

\begin{figure*}[ht]
	\centering
	\includegraphics[width=0.975\textwidth]{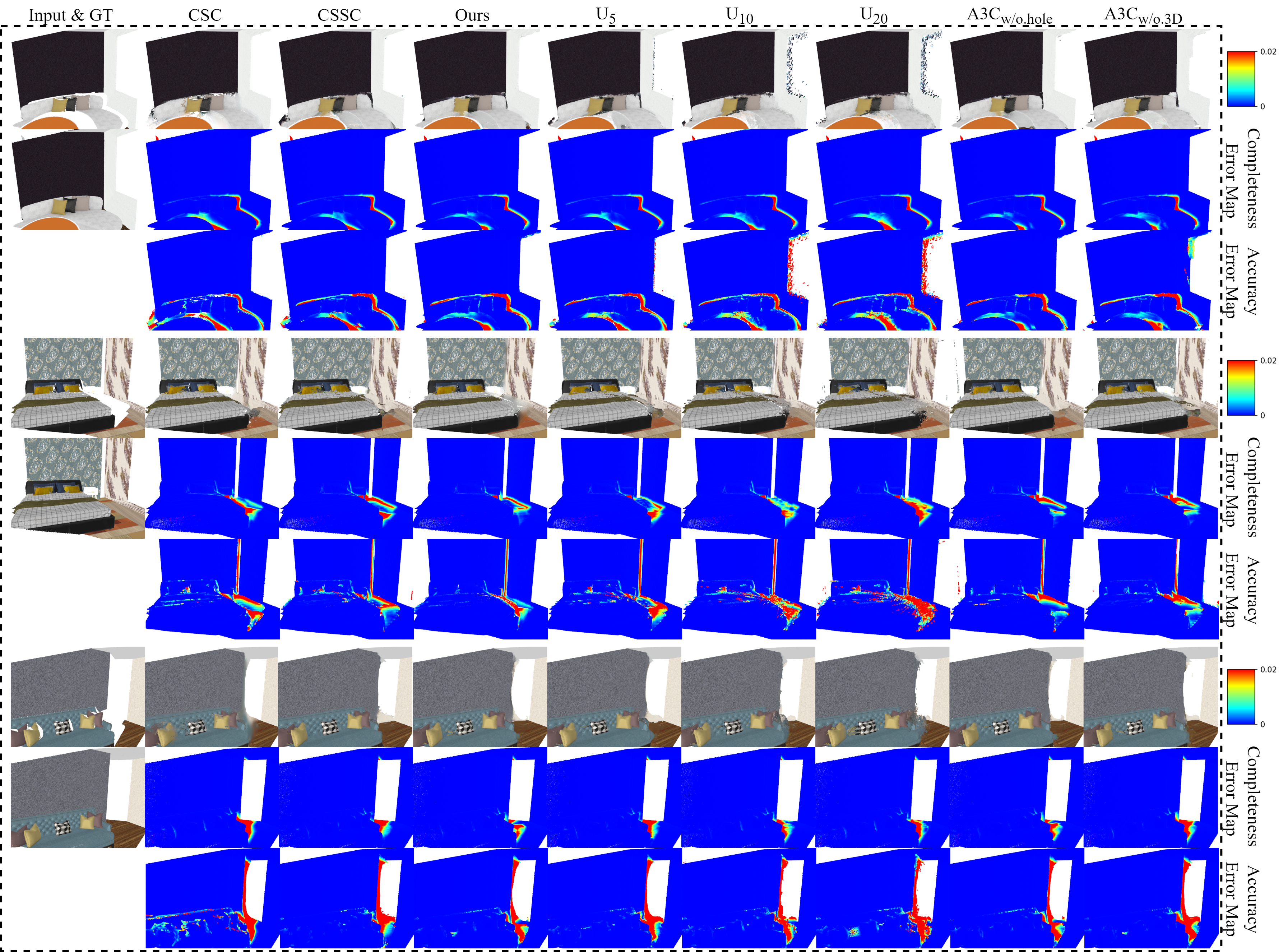}
	\caption{Comparisons on the variants of view path planning and different baselines. Given different inputs and the referenced groundtruth, we show the colored completion results of seven different approaches, {\color{zzx}with the corresponding completeness and accuracy error maps below.}}
	\label{fig:ablation}
	\vspace{-0.5cm}
\end{figure*}

\begin{figure}[ht]
	\centering
	\includegraphics[width=0.475\textwidth]{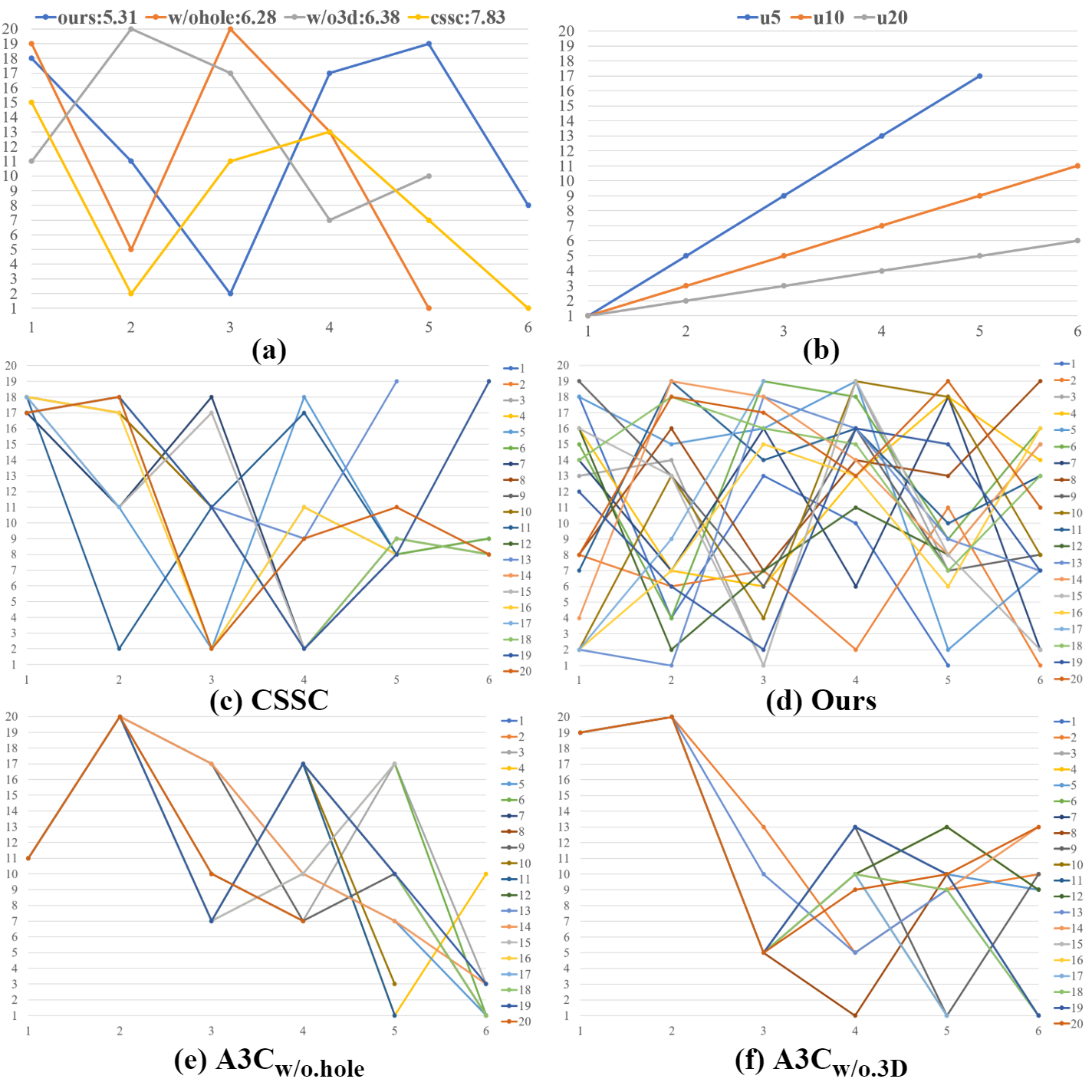}
	\caption{{\color{zzx}Ablation studies on view path planning. For all charts, the horizontal coordinate denotes the number of steps and the vertical coordinate indicates the index of the selected viewpoint. Only the first 6 viewpoint selections are shown in this figure. (a) The view selection paths for the same scene by the four reinforcement learning methods ($ A3C $, $ A3C_{w/o.hole} $, $ A3C_{w/o.3d} $ and $ DQN $). Also shown on the legend are the average numbers of selected viewpoints on the 3D-FUTURE test set. (b) The view selection paths of $ U_5 $, $ U_{10} $ and $ U_20 $. (c-f) Viewpoint selection results of different methods on the same 20 random scenes.}}
	\label{fig:abl_view_graph}
	\vspace{-0.2cm}
\end{figure}

\begin{figure}[ht]
	\centering
	\includegraphics[width=0.475\textwidth]{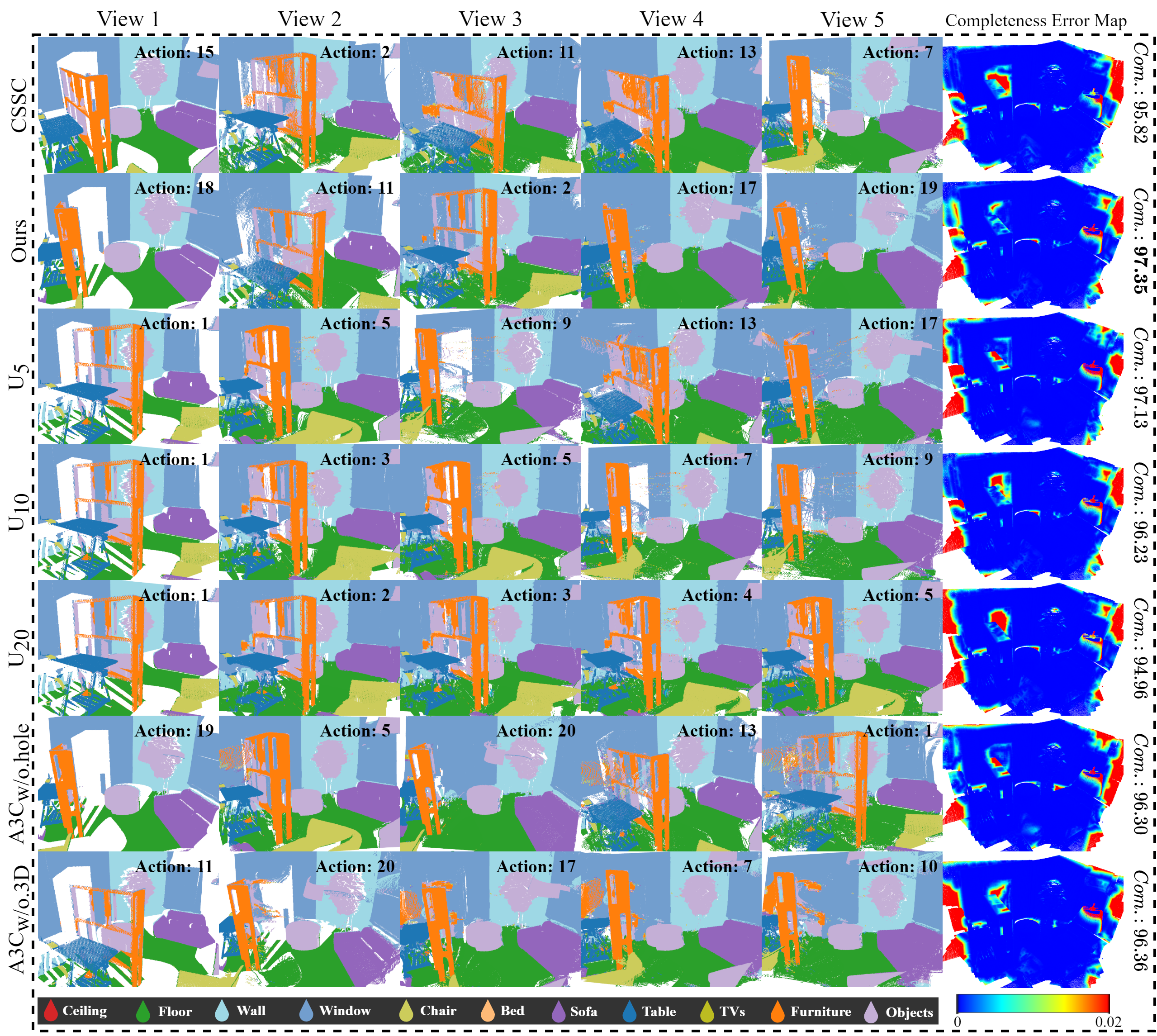}
	\caption{{\color{zzx}Visualization of different chosen viewpoint paths for several methods. The last column is the completeness error map of the corresponding reconstruction results with $ C_{0.02} $ marked on the side.} }
	\label{fig:abl_view_fig}
	\vspace{-0.5cm}
\end{figure}

Without using reinforcement learning for path planning, there exists a straightforward way to do completion: we can uniformly sample views from $C${ \color{zzx}with different steps} and directly perform image inpainting on them. In this uniform manner, three methods with three different numbers of views are evaluated. We denote them as $U_5$, $ U_{10} $ and $U_{20}$ {\color{zzx}(with steps $ 4,2 $ and $ 1 $)}. In addition, we also train a new A3C with only the reward function $ R_{i}^{acc}$, denoted as $Ours_{w/o.hole}$, and $Ours_{w/o.3D}$ without the reward function $ R_i^{pcacc} $. Visual comparison results on some sampled scenes are shown in Fig. \ref{fig:ablation} and Fig. \ref{fig:abl_view_fig}, where our proposed model results in much better appearances than others. {\color{zzx}The selected viewpoints of the first 6 steps for some randomly chosen scenes are shown in Fig. \ref{fig:abl_view_graph}, where we can find that different reinforcement learning strategies will perform different view path planning for the same scene.} The quantitative results are reported in Tab \ref{tab:pc_on_3df}, from which we observe that:\\
\bm{$ U_5 $} v.s. \bm{$ U_{10} $} v.s. \bm{$ U_{20} $} v.s. \bm{$ Ours $}: An optimal sequence of viewpoints determined by a reinforcement learning algorithm will help scene color completion more effectively than a serial way. It is not that the more viewpoints for view inpainting, the better the reconstruction quality. A reasonable RL agent can ensure the scene color completion quality as well as complete the scene faster. From Tab. \ref{tab:pc_on_3df} we can observe that: although the CD value is getting smaller as the number of viewing angles increases, $ A_r $ is also decreasing at the same time. This shows that a large number of viewpoints brings full coverage of the scene, but this also leads to an increase in noisy points which affects the generation of correct position points.

\noindent \bm{$ Ours_{w/o.hole} $} v.s. \bm{$ Ours $}: The reward function $ R_{i}^{hole} $ is efficient of scene color completion process, and we also observe that $ Ours_{w/o.hole} $ chooses $ 6.28 $ viewpoints on average since it tends to pick views with small holes for higher $R_{i}^{acc}$.

\noindent \bm{$ Ours_{w/o.3D} $} v.s. \bm{$ Ours $}: With the reward function $ R_{i}^{pcacc} $, the CD is decreased compared to the one without the function, and the completed points will look closer to the surface of the real scene.

\begin{table}[htb]	
	\centering
	\caption{ {\color{zzx}Quantitative ablation studies on different action space settings.} \\ Case $ a $: $ \theta \in \{ 60^{\circ},80^{\circ} \} $ , $ \phi \in \{-50^{\circ},-40^{\circ},..., \-10^{\circ}, 10^{\circ},...,50^{\circ} \} $. \\ Case $ b $: $ \theta \in \{ 70^{\circ},90^{\circ} \} $ , $ \phi \in \{-25^{\circ},-20^{\circ},..., \-5^{\circ}, 5^{\circ},...,25^{\circ} \} $. \\ Case $ c $: $ \theta \in \{ 60^{\circ},80^{\circ} \} $ , $ \phi \in \{-25^{\circ},-20^{\circ},..., \-5^{\circ}, 5^{\circ},...,25^{\circ} \} $.}
	\label{tab:actionspace}
	\begin{tabular}{c|c|ccccc}
		
		\hline
		
		& $ CD $ & $ C_{r=0.02} $ & $ C_{0.04} $ & $ C_{0.06} $ & $ C_{0.09} $ & $ C_{0.10} $  \\
		\hline
		\hline		
		$ Ours $ & \textbf{0.4095} & 79.78 & \textbf{83.65} & \textbf{85.87} & \textbf{87.39} & \textbf{88.55}  \\
		
		Case $ a $ & 0.4126 & \textbf{79.82} & 82.30 & 85.96 & 87.38 & 88.12   \\
		
		Case $ b $ & 0.5231 & 73.54 & 79.21 & 84.68 & 85.43 & 87.70\\
		
		Case $ c $ & 0.5187 & 74.63 & 78.49 & 83.62 & 86.38 & 86.61\\
		\hline
	\end{tabular}
\end{table}

\noindent \bm{$ CSSC $} v.s. \bm{$ Ours $}: For complex problems like viewpoint path planning, reinforcement learning algorithms that consider both actions and policy ($ A3C $) will be superior to action-based methods ($ DQN $). {\color{zzx}This is demonstrated in the following aspects: 1) Shorter training time. Training the DQN in $ CSSC $ took about 5 days and only 3 days to train the A3C. 2) Smaller selected viewpoints on average. As shown in Fig. \ref{fig:abl_view_graph}(a), the average number of viewpoints chosen by $ CSSC $ is $ 7.83 $ and $ 5.31 $ of ours. 3) More diverse viewpoint options. As illustrated in Fig. \ref{fig:abl_view_graph}(c,d), $ A3C $ will choose different start points for different scenes, while $ DQN $ basically focused the start point around $ v_{17} $ and $ v_{18} $. 4) Better reconstruction performance. Whether for the $ CD,C_r $ and $ A_r $ metrics in Tab. \ref{tab:pc_on_3df} or for the $ Comp. $ and semantic $ IoU $ in Tab. \ref{tab:vox_on_3df}, our approach outperforms $ CSSC $ almost across the board.}

\subsection{Limitations}

{\color{zzx}Although our method have a good reconstruction performance, however, as shown in Tab. \ref{tab:param}, the biggest problem is the long inference time. Since our method has to render 2D segmentation and RGB-D images from the point cloud multiple times, a significant chunk of the running time is spent on rendering. Therefore using a faster rendering tool such as Blender can speed up our approach. The large number of parameter size is due to the fact that our method consists of multiple 2D image processing networks. A possible solution is to consider model lightweighting. Our method can currently only complete objects within the viewing frustum. In the future, we will consider segmenting the objects at the boundary of the frustum and performing object-level completion to get a more complete and reasonable model.}

\begin{table}[htb]	
	\centering
	\caption{ {\color{zzx}The parameter size, running GPU memory and the inference time against the state-of-the-arts.}}
	\label{tab:param}
	\begin{tabular}{c|c|c|c}
		
		& \makecell[c]{$Param.$ \\ (MB)} & \makecell[c]{GPU $Mem.$ \\ (MB)} & \makecell[c]{$Inf. time$ \\ ($ s $)}  \\
		\hline
		\hline		
		$ SSCNet $ & 3.7 & 420 & 0.56  \\
		
		$ VVNet $ & 42.9 & 1,600 & 0.73   \\
		
		$ ForkNet $ & 26.2 & 583 & 0.81 \\
		
		$ Ours $ & 1,264.8 & 8,107 & 99.61 \\
		\hline
	\end{tabular}
\end{table}

\section{Conclusion}

In this paper, we propose the first surface-generated approach for colored semantic point cloud scene completion from a single RGB-D image. The missing 3D points are inferred by conducting completion on multi-view depth maps and its missing color and semantic labels by RGB and segmentation maps. To guarantee a more accurate and consistent output, a volume-guided view image inpianting network is proposed. In addition, a deep reinforcement learning framework is devised to seek the optimal view path to contribute the best result in accuracy. The experiments demonstrate that our model is the best choice and significantly outperforms existing methods. There is a research direction worth further exploration in the future: how to post-process the generated point cloud to guarantee the completion is watertight.


%

%



\ifCLASSOPTIONcaptionsoff
  \newpage
\fi



\bibliographystyle{IEEEtran}
\bibliography{rgbdscecomp}
%
%
%

%
\vspace{3cm}
\begin{IEEEbiography}[{\includegraphics[width=1in,height=1.25in,clip,keepaspectratio]{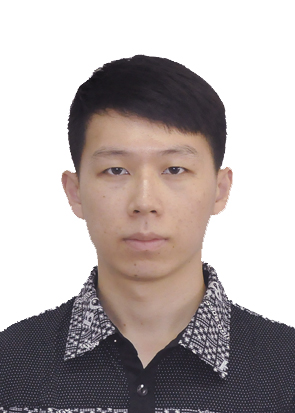}}]{Zhaoxuan Zhang}
received the B.E. degree in School of Mathematical Sciences from Dalian University of Technology, Liaoning, China in 2016, where he is currently pursuing the Ph.D. degree in computer science. His current research interests include computer vision and computer graphics, especially the 3D reconstruction.
\end{IEEEbiography}

\vspace{-10mm}

\begin{IEEEbiography}[{\includegraphics[width=1in,height=1.25in,clip,keepaspectratio]{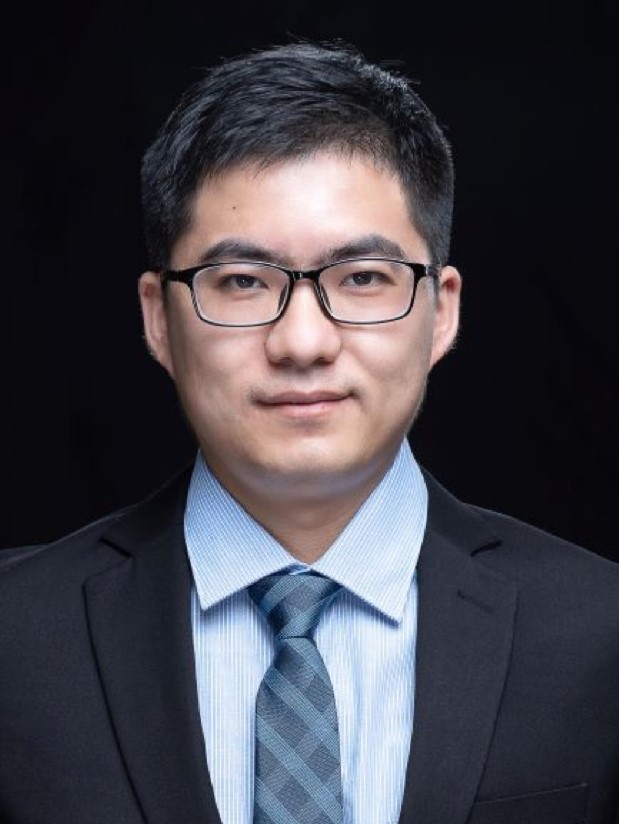}}]{Xiaoguang Han}
	is now a Research Assistant	Professor at The Chinese University of Hong	Kong, Shenzhen. He received his Ph.D. degree in computer science from The University of Hong Kong (2013-2017), his M.S. degree in applied mathematics from Zhejiang University (2009-2011) and his B.S. degree in math from Nanjing University of Aeronautics and Astronautics. He also spent 2 years (2011-2013) in City University of Hong Kong as a research associate. His research interests include computer vision, com-\\
	\textcolor{white}{xxxxxxxxxxxxxxxxxxx} puter graphics, human-computer interaction,
	\textcolor{white}{xxxxxxxxxxxxxxxxxxxx} medical image analysis and machine learning.
\end{IEEEbiography}
\vspace{-10mm}
\begin{IEEEbiography}[{\includegraphics[width=1in,height=1.25in,clip,keepaspectratio]{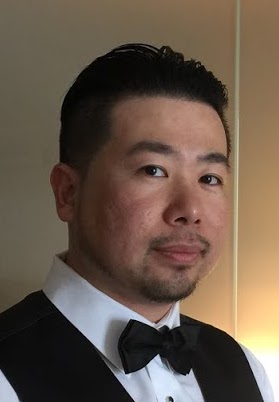}}]{Bo Dong}
	is a visiting researcher at Princeton University. He got his Ph.D. from the College of William and Mary in 2015. At the intersection of computer graphics, computational photography, and deep learning, Dr. Dong's research covers a wide range of applications in next-generation imaging systems, especially in degraded conditions. Dr. Dong got his Bachler and Master degree from Northeastern University (China) and Texas A\&M University-Corpus Christi in 2004 and 2009, respectively.
\end{IEEEbiography}
\vspace{-10mm}
\begin{IEEEbiography}[{\includegraphics[width=1in,height=1.25in,clip,keepaspectratio]{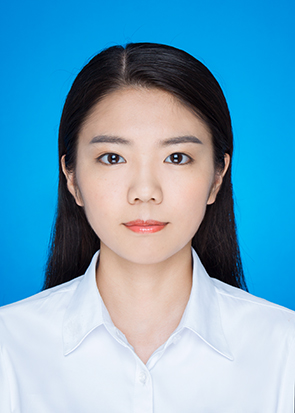}}]{Tong Li}
	
	received the B.E. degree in computer science from Wuhan University of Technology, Hubei, China in 2019. Now she is currently pursuing the M.S. degree of computer science in Dalian University of Technology. Her current research interests include computer vision, especially the 3D reconstruction and point cloud processing.
\end{IEEEbiography}
\vspace{-10mm}

\begin{IEEEbiography}[{\includegraphics[width=1in,height=1.25in,clip,keepaspectratio]{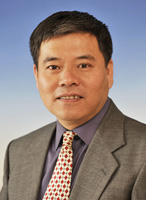}}]{Baocai Yin}
	is a professor and doctoral	advisor at Dalian University of Technology.	He received his Ph.D. degree in computational mathematics from Dalian University of Technology (1990-1993), where he also received his M.S. degree in computational mathematics (1985-1988) and his B.S. degree in applied mathematics(1981-1985). His research areas include digital multimedia technology, virtual reality and graphics technology, and multi-function perception technology.
\end{IEEEbiography}
\vspace{-10mm}
\begin{IEEEbiography}[{\includegraphics[width=1in,height=1.25in,clip,keepaspectratio]{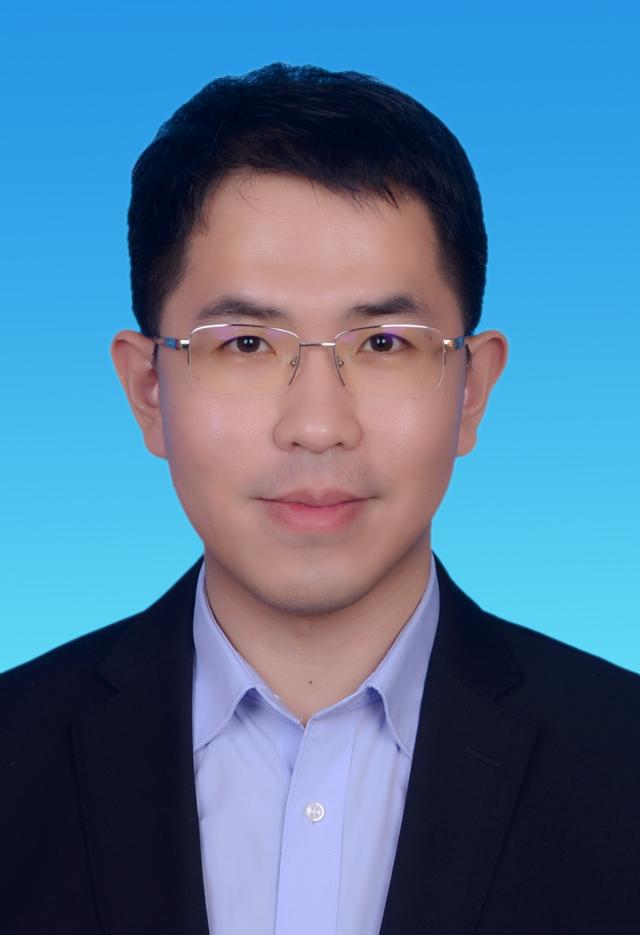}}]{Xin Yang}
	is a professor and doctoral	advisor at Dalian University of Technology.	He received his Ph.D. degree in computer science from Zhejiang University (2007-2012), and his B.S. degree in computer science from Jilin University (2003-2007). His main research interests include computer graphics and vision, intelligent	robot technology, focusing on the efficient expression, understanding, perception and interaction of scenes.
\end{IEEEbiography}

\vfill
%
%
%




\end{document}